\documentclass[runningheads]{llncs}

\usepackage[mobile]{eccv}

\usepackage{eccvabbrv}

\usepackage{graphicx}
\usepackage{booktabs}

\usepackage[accsupp]{axessibility}  

\usepackage{hyperref}

\usepackage{orcidlink}

\usepackage{longtable}
\usepackage{tabularx}

\usepackage[most]{tcolorbox}

\usepackage{alltt}
\tcbuselibrary{listingsutf8}
\usepackage[utf8]{inputenc} 
\usepackage[T1]{fontenc}    
\usepackage{url}            
\usepackage{booktabs}       
\usepackage{amsfonts}       
\usepackage{nicefrac}       
\usepackage{microtype}      
\usepackage{xcolor}         

\usepackage{lineno}
\usepackage{graphicx}
\usepackage{multirow}
\usepackage{adjustbox}
\usepackage{makecell}

\usepackage{algorithm}
\usepackage{algpseudocode}
\usepackage{algorithmicx}
\usepackage{amsmath}

\usepackage{wrapfig}
\usepackage{enumitem}

\usepackage{listings}

\usepackage{caption}

\begin{document}

\title{StarDojo: Benchmarking Open-Ended Behaviors of Agentic Multimodal LLMs in Production–Living Simulations with Stardew Valley}

\titlerunning{StarDojo}




\author{
Weihao Tan\inst{1}\thanks{Equal contribution}\orcidlink{0000-0002-5888-6081} \and
Changjiu Jiang\inst{2}\textsuperscript{$\star$}\orcidlink{0009-0008-4629-2391} \and
Yu Duan\inst{1}\textsuperscript{$\star$}\orcidlink{0009-0000-2262-3362} \and
Mingcong Lei\inst{3}\orcidlink{0009-0004-2980-1810} \and
Jiageng Li\inst{1}\orcidlink{0009-0005-4488-771X} \and
Yitian Hong\inst{4}\orcidlink{0009-0004-9551-2638} \and
Xinrun Wang\inst{5}\thanks{Corresponding author}\orcidlink{0000-0003-3369-219X} \and
Bo An\inst{1}\orcidlink{0000-0002-7064-7438}
}

\authorrunning{W.~Tan et al.}

\institute{
Nanyang Technological University, Singapore
\and
Harbin Institute of Technology, Shenzhen, China
\and
The Chinese University of Hong Kong, Shenzhen, China
\and
East China University of Science and Technology, China
\and
Singapore Management University, Singapore\\
\email{weihao001@ntu.edu.sg} \\
\url{https://weihaotan.github.io/StarDojo}
}

\maketitle

\begin{abstract}
    Autonomous agents navigating human society must master both production activities and social interactions, yet existing benchmarks rarely evaluate these skills simultaneously. To bridge this gap, we introduce StarDojo, a novel benchmark based on Stardew Valley, designed to assess AI agents in open-ended production–living simulations. In StarDojo, agents are tasked to perform essential livelihood activities such as farming and crafting, while simultaneously engaging in social interactions to establish relationships within a vibrant community. StarDojo features 1,000 meticulously curated tasks across five key domains: farming, crafting, exploration, combat, and social interactions. Additionally, we provide a compact subset of 100 representative tasks for efficient model evaluation. The benchmark offers a unified, user-friendly interface that eliminates the need for keyboard and mouse control, supports all major operating systems, and enables the parallel execution of multiple environment instances, making it particularly well-suited for evaluating the most capable foundation agents, powered by multimodal large language models (MLLMs). Extensive evaluations of state-of-the-art MLLMs agents demonstrate substantial limitations, with the best-performing model, GPT-4.1, achieving only a 12.7\% success rate, primarily due to challenges in visual understanding, multimodal reasoning and low-level manipulation. As a user-friendly environment and benchmark, StarDojo aims to facilitate further research towards robust, open-ended agents in complex production-living environments. 
\end{abstract}

\section{Introduction}
\label{sec:intro}
The complexity of human society is embodied in ``\textbf{production-living systems}''~\cite{fu2023spatial}, where individuals simultaneously engage in productive activities (e.g., farming, crafting, resource generation) and living activities (e.g., social interaction, relationship building, cultural participation).
These activities are not independent processes but mutually reinforcing and constraining components of daily life. Productive outputs provide the material foundation that enables social participation, while social relationships, community engagement, and economic exchanges directly influence access to resources, knowledge, and opportunities for production. Time, energy, and environmental dynamics further bind these processes together, forcing individuals to continuously balance resource generation, social commitments, and long-term planning under shared constraints.


While recent models have demonstrated strong performance on well-defined tasks such as question answering~\cite{achiam2023gpt}, code synthesis~\cite{li2022competition}, and mathematical problem solving~\cite{trinh2024solving}, developing human-level agentic multimodal LLMs (MLLMs) capable of operating within interactive production-living systems poses a fundamentally different challenge. Traditional interactive benchmarks~\cite{bellemare2013arcade, vinyals2019grandmaster, hafner2021benchmarking, fan2022minedojo} largely emphasize either skill execution and strategic gameplay, where textual information plays only a minimal role, and socially situated interactions are either absent or highly simplified. Other social deduction benchmarks~\cite{park2023generative, light2023avalonbench, agarwal2025wolf} focus on multi-agent reasoning communication, without structured production activities or resource-generation mechanisms. As a result, social interactions in these environments are largely decoupled from material constraints and long-term developmental trade-offs. As a result, the field still lacks a systematic assessment of an agent’s ability to operate within the coupled, constraint-bound structures that characterize human life.

\begin{figure*}[t]
    \centering
    \includegraphics[width=1\linewidth]{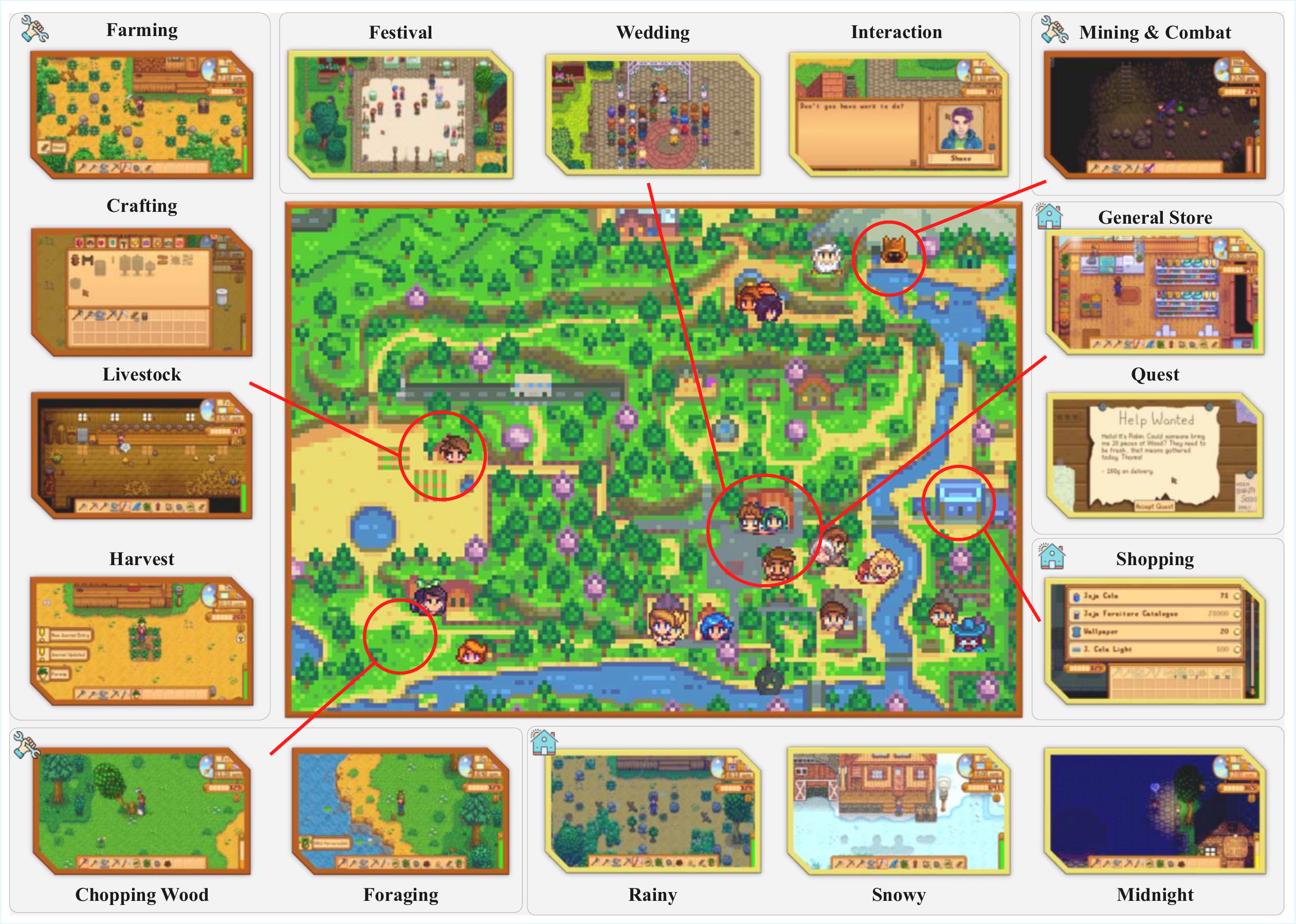}
    \caption{StarDojo leverages the open-world richness of Stardew Valley to provide a diverse array of scenarios and activities. Agents are required not only to engage in a wide range of production activities, such as farming, crafting, mining, logging, and animal husbandry, but also to participate in various social events, including trading, conversing with NPCs, and even starting families. In addition, agents need to adapt to dynamic changes in time and weather, thereby reflecting the multifaceted nature of real-world living and social interaction.  }
    \label{fig:game_play_demo}
\end{figure*}

To bridge the gap, we introduce StarDojo, a novel environment and benchmark consisting of 1,000 comprehensive tasks based on the production-living dynamics of the popular simulation game, Stardew Valley. As illustrated in Figure~\ref{fig:game_play_demo}, agents undertake tasks such as clearing farmland, tilling soil, planting and watering crops, harvesting produce, raising animals, mining and foraging for resources, and crafting essential tools and equipment.

Additionally, agents must explore diverse maps, collect various items, combat monsters, trade with merchants to generate income, upgrade and expand their farms, and complete numerous in-game quests. Social progress involves participating in festivals and community events, building interpersonal relationships, and may eventually lead to marriage and raising children. Beyond these, agents must explore diverse maps, collect various items, combat monsters, trade with merchants to generate income, upgrade and expand their farms, and complete numerous in-game quests.

Crucially, production and social progress form a bidirectional feedback loop. On one hand, social progression can unlock new recipes, discounts, services, and collaboration opportunities that directly boost production efficiency (e.g., gaining access to advanced crafting recipes or other production, enhancing resources through relationships and community participation). On the other hand, production enables social advancement by generating gifts and valuable items, providing income and materials for participating in festivals and community events, and supporting longer-term milestones such as marriage and raising children. Social progress involves engaging in festivals and community events, building interpersonal relationships, and may eventually lead to marriage and family life, each of which can further reshape production possibilities through newly available items, rewards, and opportunities.

The environment realistically simulates time, stamina, weather patterns, and seasonal cycles, all of which significantly impact both production tasks and social interactions, presenting further challenges for agents. To facilitate development and evaluation for researchers, we also present StarDojo-Lite, a curated subset comprising 100 core tasks that focus on the essential skills typically encountered during the game's early stages.

To facilitate the development and evaluation of agent behaviors, StarDojo offers four key features:
1) \textbf{Unified User-friendly Interface}: Provides an intuitive Python interface to interact seamlessly with Stardew Valley’s game engine implemented in C\#, simplifying interaction and internal state capture, eliminating the need for manual screenshots and keyboard/mouse inputs to obtain observation and action.
2) \textbf{Automated Evaluation}. Includes comprehensive evaluation scripts for all tasks, ensuring reliable and reproducible agent performance assessments.
3) \textbf{System Compatibility}. Supports major operating systems (Ubuntu, macOS, and Windows) to ensure wide accessibility.
4) \textbf{Parallelized Environments}.
Allows multiple headless environment instances to run concurrently, significantly enhancing efficiency for evaluation and data collection.

Through extensive evaluations, we demonstrate that tasks within StarDojo present significant challenges even for agents with state-of-the-art MLLMs. Our assessments cover cutting-edge models, including GPT-4.1 (\& mini)~\cite{openai2025gpt4.1}, Claude-3.7 Sonnet~\cite{anthropic2025claude3.7}, Gemini 2.5 Pro~\cite{google2025gemini2.5}, and the open-source Llama 4 Maverick~\cite{meta2025llama4}, Qwen2.5-VL-72B~\cite{bai2025qwen2} and Gemma 3 27B~\cite{team2025gemma}. These agents achieve performance ranging from 4\% to 12.7\% on the StarDojo-Lite task suite.
While agents successfully complete some easy-level tasks requiring fewer than 30 steps, they exhibit near-zero success rates on medium and hard tasks demanding extended action sequences. We conducted a detailed error analysis and attributed these failures to four primary categories: visual understanding, multimodal reasoning, long-term planning, and low-level control.
We release StarDojo as an open-source environment and benchmark, providing setup instructions, robust evaluation scripts, comprehensive documentation, and baseline implementations. We hope it can facilitate research into robust, open-ended decision-making agents capable of lifelong learning and long-term planning in production-living systems.

\begin{table*}[ht]
    \caption{Comparison of StarDojo with representative existing environments. The columns indicate whether the environment supports open-ended interaction and continuous learning, evaluates the ability to perform long-term planning, reflects scenarios relevant to real-world daily life, includes production activities (e.g., farming, hunting, and crafting), simulates human society with social interactions, implements a realistic economy supporting production and social activities, and provides language-based APIs for interaction with LLMs.}
    \label{tab:related_work}
    \small
    \setlength{\tabcolsep}{1pt} 
    \centering
    \begin{adjustbox}{max width=\textwidth}
    \begin{tabular}{c|ccccccc}
    \hline
               Environment & \begin{tabular}[c]{@{}c@{}}Open-\\ Endness\end{tabular}                   & \begin{tabular}[c]{@{}c@{}}Long-Term\\ Planning\end{tabular}                      & \begin{tabular}[c]{@{}c@{}}Routine\\ Planning\end{tabular}                              & \begin{tabular}[c]{@{}c@{}}Production\\ Activities\end{tabular}                                                                        & \begin{tabular}[c]{@{}c@{}}Social\\ Interaction\end{tabular}                       & \begin{tabular}[c]{@{}c@{}}Economy\\ System\end{tabular}                          & \begin{tabular}[c]{@{}c@{}}Language\\ APIs\end{tabular}                           \\ \hline
    Atari~\cite{bellemare2013arcade}      & \scalebox{0.85}[1]{$\times$}                               & \scalebox{0.85}[1]{$\times$}                               & \scalebox{0.85}[1]{$\times$}                               & \scalebox{0.85}[1]{$\times$}                               & \scalebox{0.85}[1]{$\times$}                               & \scalebox{0.85}[1]{$\times$}                               & \scalebox{0.85}[1]{$\times$}                               \\
    VirtualHome~\cite{puig2018virtualhome}    & \scalebox{0.85}[1]{$\times$}                               & \scalebox{0.85}[1]{$\times$}                               & \raisebox{0.6ex}{\scalebox{0.7}{$\sqrt{}$}} & \scalebox{0.85}[1]{$\times$}                               & \scalebox{0.85}[1]{$\times$}                               & \scalebox{0.85}[1]{$\times$}                               & \scalebox{0.85}[1]{$\times$}                               \\
    SMAC~\cite{samvelyan2019starcraft}       & \scalebox{0.85}[1]{$\times$}                               & \scalebox{0.85}[1]{$\times$}                               & \scalebox{0.85}[1]{$\times$}                               & \scalebox{0.85}[1]{$\times$}                               & \scalebox{0.85}[1]{$\times$}                               & \scalebox{0.85}[1]{$\times$}                               & \scalebox{0.85}[1]{$\times$}                               \\
    Crafter~\cite{hafner2021benchmarking}   & \raisebox{0.6ex}{\scalebox{0.7}{$\sqrt{}$}} & \scalebox{0.85}[1]{$\times$} & \scalebox{0.85}[1]{$\times$}                               & \raisebox{0.6ex}{\scalebox{0.7}{$\sqrt{}$}} & \scalebox{0.85}[1]{$\times$}                               & \scalebox{0.85}[1]{$\times$}                               & \scalebox{0.85}[1]{$\times$} \\
    MineDojo~\cite{fan2022minedojo}   & \raisebox{0.6ex}{\scalebox{0.7}{$\sqrt{}$}} & \raisebox{0.6ex}{\scalebox{0.7}{$\sqrt{}$}} & \scalebox{0.85}[1]{$\times$}                               & \raisebox{0.6ex}{\scalebox{0.7}{$\sqrt{}$}} & \scalebox{0.85}[1]{$\times$}                               & \scalebox{0.85}[1]{$\times$}                               & \raisebox{0.6ex}{\scalebox{0.7}{$\sqrt{}$}} \\
    OSWorld~\cite{xie2025osworld}    & \scalebox{0.85}[1]{$\times$}                               & \raisebox{0.6ex}{\scalebox{0.7}{$\sqrt{}$}} & \raisebox{0.6ex}{\scalebox{0.7}{$\sqrt{}$}} & \scalebox{0.85}[1]{$\times$}                               & \scalebox{0.85}[1]{$\times$}                               & \scalebox{0.85}[1]{$\times$}                               & \raisebox{0.6ex}{\scalebox{0.7}{$\sqrt{}$}} \\
    TextWorld~\cite{cote2019textworld}  & \scalebox{0.85}[1]{$\times$}                               & \scalebox{0.85}[1]{$\times$}                               & \raisebox{0.6ex}{\scalebox{0.7}{$\sqrt{}$}} & \scalebox{0.85}[1]{$\times$}                               & \scalebox{0.85}[1]{$\times$}                               & \scalebox{0.85}[1]{$\times$}                               & \raisebox{0.6ex}{\scalebox{0.7}{$\sqrt{}$}}                               \\
    AvalonBench~\cite{light2023avalonbench} & \scalebox{0.85}[1]{$\times$} & \raisebox{0.6ex}{\scalebox{0.7}{$\sqrt{}$}} & \scalebox{0.85}[1]{$\times$}                               & \scalebox{0.85}[1]{$\times$}                               & \raisebox{0.6ex}{\scalebox{0.7}{$\sqrt{}$}}                               & \scalebox{0.85}[1]{$\times$}                               & \raisebox{0.6ex}{\scalebox{0.7}{$\sqrt{}$}} \\
    Smallville~\cite{park2023generative} & \raisebox{0.6ex}{\scalebox{0.7}{$\sqrt{}$}} & \raisebox{0.6ex}{\scalebox{0.7}{$\sqrt{}$}} & \raisebox{0.6ex}{\scalebox{0.7}{$\sqrt{}$}} & \scalebox{0.85}[1]{$\times$}                               & \raisebox{0.6ex}{\scalebox{0.7}{$\sqrt{}$}} & \scalebox{0.85}[1]{$\times$}                               & \raisebox{0.6ex}{\scalebox{0.7}{$\sqrt{}$}} \\
    CivRealm~\cite{qi2024civrealm}   & \raisebox{0.6ex}{\scalebox{0.7}{$\sqrt{}$}} & \raisebox{0.6ex}{\scalebox{0.7}{$\sqrt{}$}} & \scalebox{0.85}[1]{$\times$}                               & \raisebox{0.6ex}{\scalebox{0.7}{$\sqrt{}$}} & \raisebox{0.6ex}{\scalebox{0.7}{$\sqrt{}$}} & \raisebox{0.6ex}{\scalebox{0.7}{$\sqrt{}$}} & \raisebox{0.6ex}{\scalebox{0.7}{$\sqrt{}$}} \\ \hline
    \textbf{StarDojo}   & \raisebox{0.6ex}{\scalebox{0.7}{$\sqrt{}$}} & \raisebox{0.6ex}{\scalebox{0.7}{$\sqrt{}$}} & \raisebox{0.6ex}{\scalebox{0.7}{$\sqrt{}$}} & \raisebox{0.6ex}{\scalebox{0.7}{$\sqrt{}$}} & \raisebox{0.6ex}{\scalebox{0.7}{$\sqrt{}$}} & \raisebox{0.6ex}{\scalebox{0.7}{$\sqrt{}$}} & \raisebox{0.6ex}{\scalebox{0.7}{$\sqrt{}$}} \\ \hline
    \end{tabular}
    \end{adjustbox}
\end{table*}

\section{Related Work}

\textbf{Interactive Benchmarks.}
The development of robust benchmarks for evaluating decision-making agents across various scenarios has been a critical focus in AI research. Traditional reinforcement learning (RL) benchmarks~\cite{bellemare2013arcade, todorov2012mujoco, guss2019minerl, samvelyan2019starcraft} predominantly emphasize low-level control tasks and non-realistic environments.  More realistic simulators~\cite{chang2017matterport3d, kolve2017ai2, li2021igibson, puig2023habitat, puig2018virtualhome}, usually focus on embodied tasks in household scenarios, lacking significant environmental dynamics and diverse social activities. 
Recent advancements in generative agents have enabled large language models (LLMs) to simulate human social behaviors in interactive environments \cite{park2023generative, light2023avalonbench, agarwal2025wolf, cote2019textworld}. Their limited action spaces, primarily restricted to dialogue, hinder engagement in broader, real-world-inspired tasks that involve production, consumption, and resource management. On the other side, while GUI benchmarks~\cite{shi2017world, zhou2023webarena, koh2024visualwebarena, xie2025osworld} also provide interactive environments, they focus on short-term software manipulations. The most relevant work is MineDojo~\cite{fan2022minedojo}, which offers a diverse range of tasks within the open-ended environment of Minecraft. However, its gameplay primarily focuses on interactions with nature, with limited opportunities for human-like social interactions. Additionally, its complex 3D navigation controls pose significant challenges, particularly for LLM-based agents to complete even simple tasks, further limiting its suitability for more complex activities. 
Another relevant benchmark, CivRealm~\cite{qi2024civrealm},  evaluates agents' strategic decision-making at a country level within a turn-based, Civilization-like game. While CivRealm presents a variety of tasks, including managing population, production, and economy, its scope remains at a macro-strategic level, distinct from StarDojo’s focus on granular, individual-level decision-making. Additionally, recent works~\cite{paglieri2024balrog, zheng2025v} tend to evaluate MLLMs on traditionally RL environments in simple game scenarios, lacking semantic richness and social interaction. 


As shown in Table~\ref{tab:related_work}, StarDojo addresses the limitations of prior benchmarks by providing a comprehensive, open-ended evaluation platform for decision-making agents in a dynamic environment. Although StarDojo, like most of the aforementioned benchmarks, adopts an abstract visual style, it effectively reconstructs and abstracts real-world events and social dynamics in a structured and operational manner. Its unique combination of complexity, realism, and diversity makes it a valuable testbed for advancing agents in real-world-like environments.

\textbf{MLLMs Agents.}
Traditional reinforcement learning (RL) agents ~\cite{mnih2015human, lillicrap2015continuous, schulman2017proximal, haarnoja2018soft} primarily focus on low-level control and fail to leverage natural language understanding, making them unsuitable for tasks requiring complex reasoning, long-term planning, and social interactions. Recent advancements in LLM-based agents have significantly expanded AI capabilities by integrating reasoning mechanisms such as chain-of-thought (CoT) prompting~\cite{yao2023react} and reflection~\cite{shinn2023reflexion}. Modular frameworks and multi-agent architectures have enabled LLM-based agents to achieve remarkable performance in tasks like code generation~\cite{hong2023metagpt, wu2023autogen, wang2024openhands} and GUI manipulation~\cite{zheng2023synapse, zhang2024ufo, wu2024copilot, wang2024mobile}.
Additionally, Voyager~\cite{wang2023voyager} has demonstrated strong in-context lifelong learning abilities, showcasing exceptional proficiency in the open-ended world of Minecraft. However, Voyager's strong reliance on built-in APIs makes it challenging to adapt to other games.
Cradle~\cite{tan2024cradle} successfully completes meaningful tasks across multiple commercial video games and software applications with a unified interface without the need to access APIs. Its preliminary experiments in Stardew Valley highlight the limitations of current state-of-the-art agents, particularly in handling multi-modal understanding, long-term planning, and resource management, which reveals the importance of extending Stardew Valley as a well-developed benchmark for decision-making agents.

\begin{figure*}[t]
    \centering
    \includegraphics[width=\linewidth]{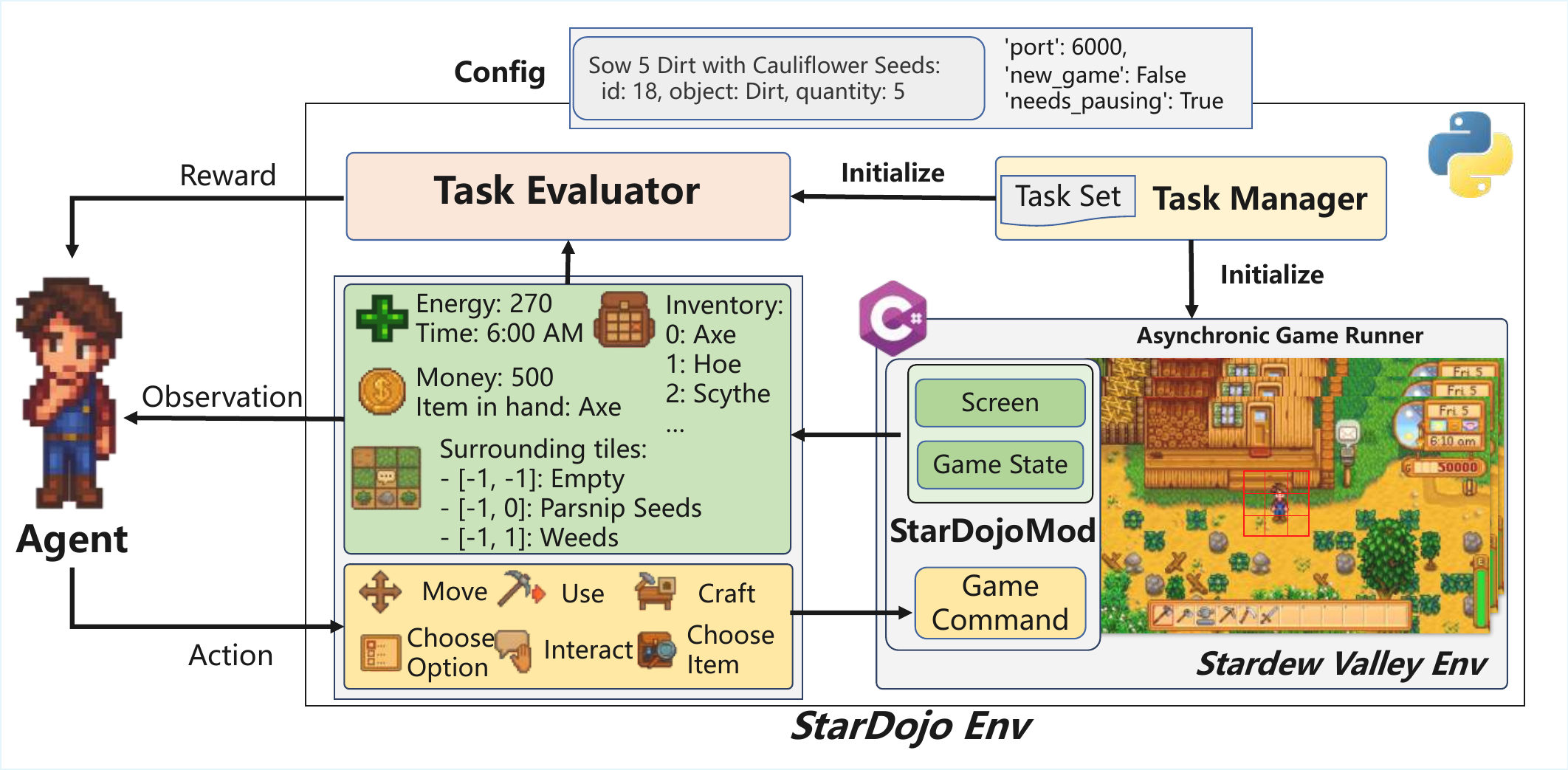}
    \caption{StarDojo environment is initiated by configurable task files. It communicates with parallel game engines through StarDojoMod to obtain internal game states and execute commands, which will be encapsulated as observations and actions by the Python Wrapper.}
    \label{fig:architecture}
\end{figure*}

\section{StarDojo}
\subsection{Introduction to Stardew Valley}
Stardew Valley is a globally popular open-ended simulation RPG game where players inherit a run-down farm. Players must thoughtfully manage their farming strategies, explore the surrounding village, build meaningful relationships with villagers,  and gather diverse resources to revitalize the farm. More details can be found in Appendix~\ref{appendix:intro_stardew}.

\textbf{Realistic Dynamics}. 
Each in-game day in Stardew Valley lasts from 6 AM to 2 AM. When night falls, outdoor activities are affected by the darkness. Staying awake past midnight will result in penalties. Players start each day with 270 energy points, spent through activities like farming and mining, and restored by sleeping or eating. The game features four 28-day seasons and different daily weather conditions, each affecting farming,  forage and other events. Effective management of time and energy is crucial to maximizing productivity and overcoming the game's primary challenges.

\textbf{Rich Production Activities}. 
As the core gameplay of Stardew Valley, players engage in a wide range of productive activities, including clearing land, cultivating crops, raising animals, mining, and foraging. The fruits of their labor, such as harvested crops, animal products, and crafted goods, can be sold for income, reinvested to expand and improve the farm, or given to villagers to strengthen friendships and complete quests or community objectives. 
Through these interconnected systems, players gradually enhance both the sustainability of their farm and their overall quality of life within the game world.

\textbf{Diverse Social Interaction}. 
The game also features social interaction with 45 unique non-player characters (NPCs), each with their own personality and routines. Players can build friendships and may even date, marry, and raise children with villagers. Improving friendships not only enriches the narrative experience but also facilitates production, as villagers may send useful gifts, share recipes, or offer assistance in various activities. Rich town festivals and quests provide further opportunities for community engagement and resource gathering.

\textbf{Comprehensive Economic System}. Agents must engage in strategic resource management, investment, and efficient planning to generate income from both production and social activities while adapting to seasonal demands and market conditions to ensure long-term financial success.

Overall, Stardew Valley serves as an ideal environment for decision-making agents in the production–living simulation. Its well-integrated systems of time management, resource allocation, economic planning, and social interaction provide a dynamic and complex environment that requires strategic thinking and adaptability. The game's structured yet open-ended nature makes it an excellent testbed for evaluating decision-making capabilities in simulated real-world conditions.


\subsection{Architecture}
As a typical commercial video game, Stardew Valley only supports human-like interaction, e.g., observing gameplay through screenshots and using keyboard and mouse to control. Additionally, the game window must remain active and in the foreground, significantly restricting automated gameplay and preventing simultaneous execution of multiple instances. As shown in Figure~\ref{fig:architecture}, to overcome these limitations, we introduce the carefully designed StarDojo environment, enabling efficient interaction and comprehensive evaluation of agents. 

\textbf{Unified User-friendly Interface}. 
We present StarDojoMod, a novel extension built upon the Stardew Modding API (SMAPI)\cite{SMAPI}, which is a widely adopted, open-source modding framework designed specifically for Stardew Valley. SMAPI offers developers extensive APIs that expose key game events and internal states, facilitating the creation of interactive and sophisticated mods. Based on SMAPI, StarDojoMod provides structured and efficient interactions between agents and the game environment. It communicates in real-time with the Stardew Valley game engine through a socket server, granting agents direct access to rendered gameplay images, saving the time-consuming screen captures, internal game states (such as character positions, statuses, and environmental information), and enabling diverse callable functions as action skills beyond traditional keyboard and mouse inputs. 
Moreover, we implemented a configurable pause-and-resume mechanism by directly modifying the inner states of the game, allowing the game to pause during model inference and agent planning, and resume before action execution. Inherited from SMAPI, StarDojoMod is implemented in C\# to be consistent with the game engine. To enhance ease-of-use and accessibility of the environment, we provide a user-friendly Python Wrapper based on the StarDojoMod for observation retrieval, action execution, and task customization, empowering users to engage with the StarDojo environment effortlessly.


\textbf{System Compatibility}. Stardew Valley is one of the few games that can be played on all mainstream operating systems (Linux, macOS and Windows). We also ensured the compatibility of StarDojoMod and the Python Wrapper, enabling the entire environment to run seamlessly across different systems.

\textbf{Parallel Execution}. Our architecture is designed for scalability and parallel execution. Each instance of Stardew Valley is independently managed through unique ports, enabling simultaneous control of multiple game instances without interference. Communication efficiency is further enhanced through the use of shared memory, reducing observation retrieval time to as little as 30 ms. Furthermore, StarDojoMod supports headless operation through the X Virtual Framebuffer (Xvfb), enabling compatibility with Linux systems without graphical interfaces, thus broadening accessibility across diverse hardware and system configurations.


\subsection{Observation and Action Spaces}
\textbf{Observation Space.} StarDojo offers a comprehensive observation space that integrates both visual and textual modalities to accommodate a wide range of agent architectures. Each observation includes a gameplay screenshot alongside detailed textual information describing the game state. This textual state contains character status (such as health and energy), local tile information ($n \times n$ tiles surrounding the agent), and global information (such as time, weather, and the positions of NPCs and buildings). 
By combining visual and textual observations, StarDojo ensures that agents can leverage both high-level context and fine-grained environmental cues for more robust and informed decision-making. For our experiments, we selected a subset of this information to fairly evaluate agent behaviors. Full details of the observation space are provided in Appendix~\ref{appendix:obs_space}.


\textbf{Action Space.} The action space in StarDojo is designed to encompass the full range of activities that can be performed in the original Stardew Valley using a keyboard and mouse. Actions are carefully abstracted to eliminate redundant operations while retaining the core decision-making challenges inherent to the game. We define ten fundamental actions, which together are sufficient to cover most gameplay scenarios:
\textit{move(x, y)}, 
\textit{use(direction)},  \textit{interact(direction)}, \textit{choose\_item(slot\_index)},  \textit{attach\_item(slot\_index)},  \textit{detach\_item()},  
\textit{craft(item)},  
\textit{choose\_option(option\_index, quantity, direction)}, 
\textit{menu(option, menu\_name)} and 
\textit{navigate(name)}. 
In our experiments, we excluded the \textit{navigate} action from the available action space. While \textit{navigate} provides a high-level shortcut for moving between maps, our primary objective is to evaluate the agent's realistic exploration abilities and its performance using an action space that more closely mirrors human interactions. Detailed descriptions of the action space can be found in Appendix~\ref{appendix:action_space}.

\begin{figure*}[t]
    \centering
    \begin{minipage}{0.5\linewidth}
        \centering
        \includegraphics[width=0.9\linewidth]{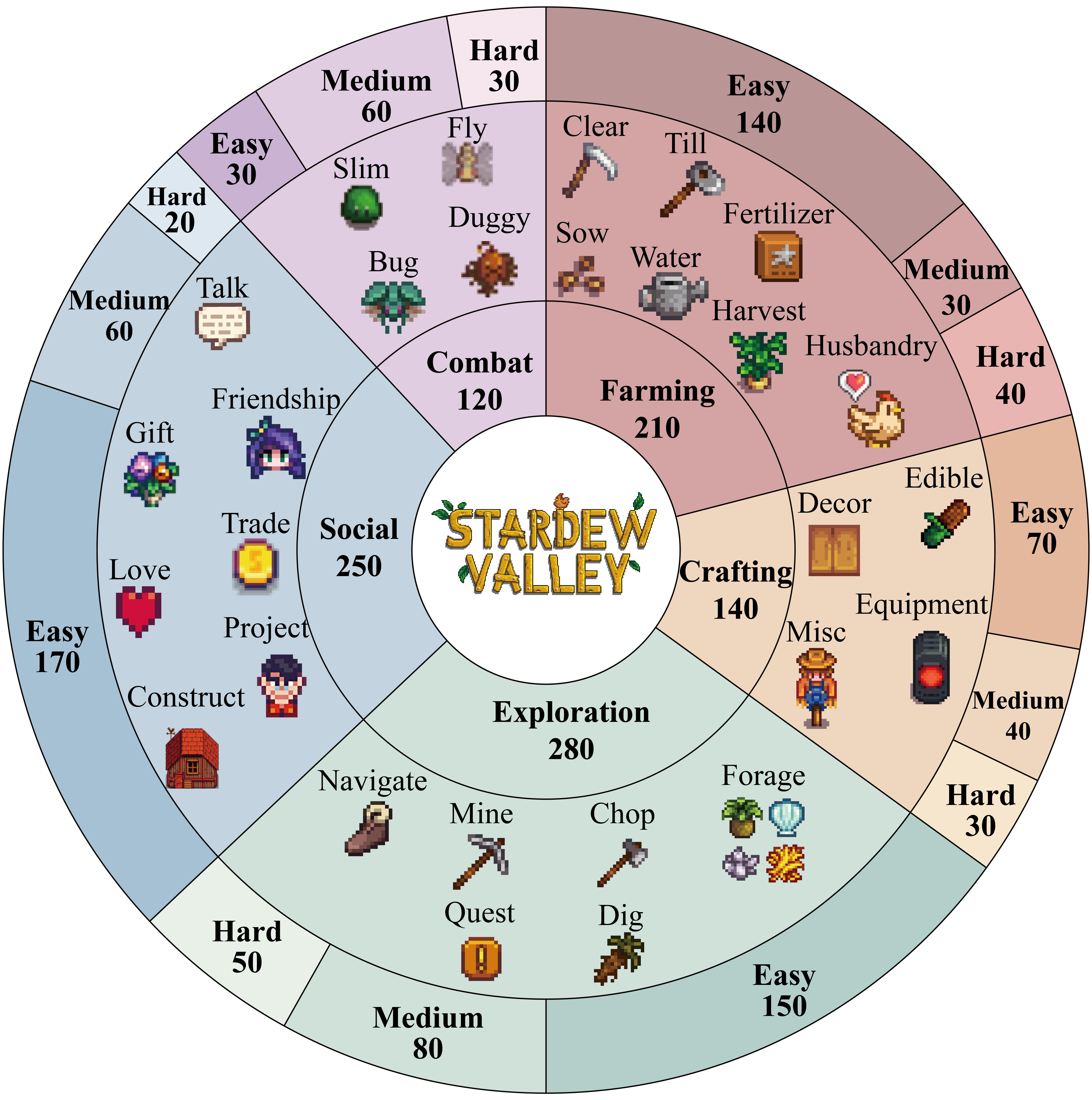}
        \caption{Distribution of 1000 tasks across five categories: Farming, Crafting, Exploration, Combat and Social in StarDojo, each with Easy, Medium, and Hard difficulties.}
        \label{fig:tasks}
    \end{minipage}%
    \hfill
    \begin{minipage}{0.45\linewidth}
        \captionof{table}{ Task statistics of StarDojo-Lite. The task suite is made up of the most representative early-stage tasks from each category.}
        \centering
        \small
        \setlength{\tabcolsep}{1pt}
        \begin{adjustbox}{width=\linewidth}
        \begin{tabular}{@{}c|cccc@{}}
        \toprule
        \textbf{Category} & \textbf{Easy} & \textbf{Medium} & \textbf{Hard} & \textbf{Total}\\
        \midrule
        Farming & 14 & 3 & 4 & 21 \\
        Crafting & 7 & 4 & 3 & 14 \\
        Exploration & 15 & 8 & 5 & 28 \\
        Combat & 3 & 6 & 3 & 12 \\
        Social & 17 & 6 & 2 & 25 \\
        Total & 56 & 23 & 21 & 100 \\
        \bottomrule
        \end{tabular}
        \end{adjustbox}
        \label{tab:lite_tasks}
    \end{minipage}
\end{figure*}

\subsection{Tasks}

As shown in Figure \ref{fig:tasks}, we carefully curate 1000 tasks to benchmark agents' various behaviors in StarDojo. These tasks are divided into five distinct categories: Farming,  Crafting, Exploration, Combat and Social, which cover most of the production-living activities in the early and middle stages of the game. Each task is classified into three difficulties: easy, medium, and hard.
For easy-level tasks,  agents are provided with all necessary items or tools from the start (e.g., mature crops ready for harvest, ingredients for crafting). Agents are initialized near the target location (e.g., inside a shop with enough budget for trading). These tasks primarily test the agent’s basic ability to complete atomic activities. For medium-level tasks, agents need to fulfill the prerequisites of the tasks on their own, such as planting and harvesting crops, gathering crafting ingredients, traveling from the farm to the target area, and earning sufficient funds to purchase specific items. Many of these resources can be acquired through multiple sources and methods. This flexibility gives agents considerable freedom in choosing their strategies. Hard-level tasks typically require several in-game days to complete and are often composed of multiple medium-level tasks. Agents have greater freedom to allocate their time and energy each day, choosing how to prioritize activities and sequence actions. Success often demands strategic long-term planning, adaptive decision-making, and efficient resource management, as there are multiple viable paths to achieving the goal.

As shown in Table~\ref{tab:lite_tasks}, to facilitate efficient agent evaluation, we curate a representative smaller task suite, called StarDojo-Lite, comprising 100 core tasks from the full task collection, balancing coverage and practicality. This lite task set covers most of the representative activities in the early stage of the game. Detailed descriptions of the task set can be found in Appendix~\ref{appendix:task}.

\textbf{Playthrough}. In addition to modular tasks, StarDojo also provides an extended Playthrough task designed to comprehensively assess an agent’s strategic decision-making over prolonged gameplay. Starting from scratch, the agent’s objective is to accumulate 1 million in-game currency, which is a significant milestone typically achieved over hundreds of in-game days. Success requires efficient seasonal planning, unlocking new maps through quests, careful resource and energy management, skill progression for advanced crafting, and strategic relationship-building with NPCs. 

\textbf{Initial Config and Setup}. Some tasks require the agent to possess specific equipment (e.g., a sword), have certain items (e.g., sufficient crop seeds in inventory), or have completed particular game progress (e.g., unlocking the mines). To establish the initial state for each task, we provide multiple saved game files reflecting various stages of progression, along with specialized task-specific functions utilizing StarDojoMod commands. At the start of each task, StarDojo automatically loads the corresponding saved game and executes these task-specific commands, ensuring all necessary prerequisites, such as items, equipment, skill levels, date and farm status, are appropriately configured.


\textbf{Automatic Evaluation}. Given the extensive number of existing tasks and the vast potential for future additions, it is essential to establish a reusable, scalable, adaptive, and efficient evaluation mechanism. StarDojo addresses this need by implementing a general evaluation system that continuously monitors task progression and provides immediate reward feedback to agents. Specifically, after agents finish executing actions at each step, the evaluator is invoked to determine whether tasks have been successfully completed or have reached the predefined maximum number of steps. To achieve this, the evaluator maintains the previous game state information, compares it with the current state obtained via StarDojoMod, identifies incremental changes indicating progress, and accurately assesses task completion criteria based on these observed differences. By leveraging a comprehensive observation space for incremental progress tracking, our evaluation approach effectively mitigates discrepancies caused by varying initial conditions, ensuring robust adaptability across diverse tasks.

\section{Empirical Studies}
In this section, we benchmark multiple advanced MLLMs on StarDojo and provide a comprehensive study and analysis of their behaviors and limitations. 

\begin{table*}[t]
  \caption{Success rates(\%) and standard deviation of agents with different base models on StarDojo-Lite task set, ranging over five categories (Farming, Crafting, Exploration, Combat and Social) and three levels of difficulty. Each task is evaluated over three runs.}
  \label{tab:agent_baseline}
  \centering
  \small
  \begin{adjustbox}{max width=\textwidth}
  \begin{tabular}{@{}c|c|cccccc@{}}
    \toprule
    \textbf{Model} & \textbf{Task} & \textbf{Farming} & \textbf{Crafting} & \textbf{Exploration} & \textbf{Combat} & \textbf{Social } & \textbf{Total} \\
    \midrule
    \multirow{4}{*}{\makecell{\textbf{GPT-4.1}}}  
      & \textbf{Easy}   & \makecell{\textbf{31.0}{\scriptsize$\pm$4.1}} & \makecell{\textbf{52.4}{\scriptsize$\pm$8.3}} & \makecell{\textbf{15.6}{\scriptsize$\pm$3.9}} & \makecell{\textbf{11.1}{\scriptsize$\pm$19.3}} & \makecell{7.8{\scriptsize$\pm$6.8}} & \makecell{\textbf{21.4}{\scriptsize$\pm$1.8}} \\
      & \textbf{Medium}   & \makecell{0.0{\scriptsize$\pm$0.0}} & \makecell{0.0{\scriptsize$\pm$0.0}} & \makecell{8.3{\scriptsize$\pm$7.2}} & \makecell{0.0{\scriptsize$\pm$0.0}} & \makecell{0.0{\scriptsize$\pm$0.0}} & \makecell{2.7{\scriptsize$\pm$2.3}} \\
      & \textbf{Hard}   & \makecell{0.0{\scriptsize$\pm$0.0}} & \makecell{0.0{\scriptsize$\pm$0.0}} & \makecell{0.0{\scriptsize$\pm$0.0}} & \makecell{0.0{\scriptsize$\pm$0.0}} & \makecell{0.0{\scriptsize$\pm$0.0}} & \makecell{0.0{\scriptsize$\pm$0.0}} \\
      & \textbf{Total}   & \makecell{\textbf{20.6}{\scriptsize$\pm$2.8}} & \makecell{\textbf{26.2}{\scriptsize$\pm$4.1}} & \makecell{\textbf{10.7}{\scriptsize$\pm$0.0}} & \makecell{\textbf{2.8}{\scriptsize$\pm$4.8}} & \makecell{5.3{\scriptsize$\pm$4.6}} & \makecell{\textbf{12.7}{\scriptsize$\pm$0.6}} \\
    \midrule
    \multirow{4}{*}{\makecell{\textbf{Gemini 2.5 Pro}}}  
      & \textbf{Easy}   & \makecell{26.2{\scriptsize$\pm$4.1}} & \makecell{47.6{\scriptsize$\pm$8.3}} & \makecell{6.7{\scriptsize$\pm$6.7}} & \makecell{0.0{\scriptsize$\pm$0.0}} & \makecell{\textbf{11.8}{\scriptsize$\pm$5.9}} & \makecell{17.9{\scriptsize$\pm$1.8}} \\
      & \textbf{Medium}   & \makecell{0.0{\scriptsize$\pm$0.0}} & \makecell{0.0{\scriptsize$\pm$0.0}} & \makecell{8.3{\scriptsize$\pm$7.2}} & \makecell{0.0{\scriptsize$\pm$0.0}} & \makecell{0.0{\scriptsize$\pm$0.0}} & \makecell{2.7{\scriptsize$\pm$2.3}} \\
      & \textbf{Hard}   & \makecell{0.0{\scriptsize$\pm$0.0}} & \makecell{0.0{\scriptsize$\pm$0.0}} & \makecell{0.0{\scriptsize$\pm$0.0}} & \makecell{0.0{\scriptsize$\pm$0.0}} & \makecell{0.0{\scriptsize$\pm$0.0}} & \makecell{0.0{\scriptsize$\pm$0.0}} \\
      & \textbf{Total}   & \makecell{17.5{\scriptsize$\pm$2.8}} & \makecell{23.8{\scriptsize$\pm$4.1}} & \makecell{6.0{\scriptsize$\pm$2.1}} & \makecell{0.0{\scriptsize$\pm$0.0}} & \makecell{\textbf{8.0}{\scriptsize$\pm$4.0}} & \makecell{10.7{\scriptsize$\pm$0.6}} \\
    \midrule
    \multirow{4}{*}{\makecell{\textbf{Claude 3.7 Sonnet}}} 
      & \textbf{Easy}   & \makecell{\textbf{31.0}{\scriptsize$\pm$4.1}} & \makecell{28.6{\scriptsize$\pm$0.0}} & \makecell{8.9{\scriptsize$\pm$10.2}} & \makecell{0.0{\scriptsize$\pm$0.0}} & \makecell{7.8{\scriptsize$\pm$3.4}} & \makecell{16.1{\scriptsize$\pm$4.7}} \\
      & \textbf{Medium}   & \makecell{0.0{\scriptsize$\pm$0.0}} & \makecell{0.0{\scriptsize$\pm$0.0}} & \makecell{\textbf{16.7}{\scriptsize$\pm$7.2}} & \makecell{0.0{\scriptsize$\pm$0.0}} & \makecell{0.0{\scriptsize$\pm$0.0}} & \makecell{\textbf{5.3}{\scriptsize$\pm$2.3}} \\
      & \textbf{Hard}   & \makecell{0.0{\scriptsize$\pm$0.0}} & \makecell{0.0{\scriptsize$\pm$0.0}} & \makecell{0.0{\scriptsize$\pm$0.0}} & \makecell{0.0{\scriptsize$\pm$0.0}} & \makecell{0.0{\scriptsize$\pm$0.0}} & \makecell{0.0{\scriptsize$\pm$0.0}} \\
      & \textbf{Total}   & \makecell{\textbf{20.6}{\scriptsize$\pm$2.8}} & \makecell{14.3{\scriptsize$\pm$0.0}} & \makecell{9.5{\scriptsize$\pm$5.5}} & \makecell{0.0{\scriptsize$\pm$0.0}} & \makecell{5.3{\scriptsize$\pm$2.3}} & \makecell{10.3{\scriptsize$\pm$2.5}} \\
    \midrule
    \multirow{4}{*}{\makecell{\textbf{GPT-4.1 mini}}} 
      & \textbf{Easy}   & \makecell{26.2{\scriptsize$\pm$4.1}} & \makecell{4.8{\scriptsize$\pm$8.3}} & \makecell{4.4{\scriptsize$\pm$7.7}} & \makecell{\textbf{11.1}{\scriptsize$\pm$19.3}} & \makecell{7.8{\scriptsize$\pm$3.4}} & \makecell{11.3{\scriptsize$\pm$2.7}} \\
      & \textbf{Medium}   & \makecell{0.0{\scriptsize$\pm$0.0}} & \makecell{0.0{\scriptsize$\pm$0.0}} & \makecell{8.3{\scriptsize$\pm$7.2}} & \makecell{0.0{\scriptsize$\pm$0.0}} & \makecell{0.0{\scriptsize$\pm$0.0}} & \makecell{2.7{\scriptsize$\pm$2.3}} \\
      & \textbf{Hard}   & \makecell{0.0{\scriptsize$\pm$0.0}} & \makecell{0.0{\scriptsize$\pm$0.0}} & \makecell{0.0{\scriptsize$\pm$0.0}} & \makecell{0.0{\scriptsize$\pm$0.0}} & \makecell{0.0{\scriptsize$\pm$0.0}} & \makecell{0.0{\scriptsize$\pm$0.0}} \\
      & \textbf{Total}   & \makecell{17.5{\scriptsize$\pm$2.8}} & \makecell{2.4{\scriptsize$\pm$4.1}} & \makecell{4.8{\scriptsize$\pm$5.5}} & \makecell{\textbf{2.8}{\scriptsize$\pm$4.8}} & \makecell{5.3{\scriptsize$\pm$2.3}} & \makecell{7.0{\scriptsize$\pm$1.7}} \\
    \midrule
    \multirow{4}{*}{\makecell{\textbf{Llama 4 Maverick}}}  
      & \textbf{Easy}   & \makecell{28.6{\scriptsize$\pm$0.0}} & \makecell{28.6{\scriptsize$\pm$0.0}} & \makecell{0.0{\scriptsize$\pm$0.0}} & \makecell{0.0{\scriptsize$\pm$0.0}} & \makecell{7.8{\scriptsize$\pm$3.4}} & \makecell{13.1{\scriptsize$\pm$1.0}} \\
      & \textbf{Medium}   & \makecell{0.0{\scriptsize$\pm$0.0}} & \makecell{0.0{\scriptsize$\pm$0.0}} & \makecell{4.2{\scriptsize$\pm$7.2}} & \makecell{0.0{\scriptsize$\pm$0.0}} & \makecell{0.0{\scriptsize$\pm$0.0}} & \makecell{1.3{\scriptsize$\pm$2.3}} \\
      & \textbf{Hard}   & \makecell{0.0{\scriptsize$\pm$0.0}} & \makecell{0.0{\scriptsize$\pm$0.0}} & \makecell{0.0{\scriptsize$\pm$0.0}} & \makecell{0.0{\scriptsize$\pm$0.0}} & \makecell{0.0{\scriptsize$\pm$0.0}} & \makecell{0.0{\scriptsize$\pm$0.0}} \\
      & \textbf{Total}   & \makecell{19.1{\scriptsize$\pm$0.0}} & \makecell{14.3{\scriptsize$\pm$0.0}} & \makecell{1.2{\scriptsize$\pm$2.1}} & \makecell{0.0{\scriptsize$\pm$0.0}} & \makecell{5.3{\scriptsize$\pm$2.3}} & \makecell{7.7{\scriptsize$\pm$0.6}} \\
    \midrule
    \multirow{4}{*}{\makecell{\textbf{Qwen2.5 VL 72B}}}  
      & \textbf{Easy}   & \makecell{16.7{\scriptsize$\pm$8.3}} & \makecell{9.5{\scriptsize$\pm$8.3}} & \makecell{13.3{\scriptsize$\pm$13.3}} & \makecell{0.0{\scriptsize$\pm$0.0}} & \makecell{9.8{\scriptsize$\pm$3.4}} & \makecell{11.9{\scriptsize$\pm$5.5}} \\
      & \textbf{Medium}   & \makecell{0.0{\scriptsize$\pm$0.0}} & \makecell{0.0{\scriptsize$\pm$0.0}} & \makecell{0.0{\scriptsize$\pm$0.0}} & \makecell{0.0{\scriptsize$\pm$0.0}} & \makecell{0.0{\scriptsize$\pm$0.0}} & \makecell{0.0{\scriptsize$\pm$0.0}} \\
      & \textbf{Hard}   & \makecell{0.0{\scriptsize$\pm$0.0}} & \makecell{0.0{\scriptsize$\pm$0.0}} & \makecell{0.0{\scriptsize$\pm$0.0}} & \makecell{0.0{\scriptsize$\pm$0.0}} & \makecell{0.0{\scriptsize$\pm$0.0}} & \makecell{0.0{\scriptsize$\pm$0.0}} \\
      & \textbf{Total}   & \makecell{11.1{\scriptsize$\pm$5.5}} & \makecell{4.8{\scriptsize$\pm$4.1}} & \makecell{7.1{\scriptsize$\pm$7.1}} & \makecell{0.0{\scriptsize$\pm$0.0}} & \makecell{6.7{\scriptsize$\pm$2.3}} & \makecell{6.7{\scriptsize$\pm$3.1}} \\
    \midrule
    \multirow{4}{*}{\makecell{\textbf{Gemma 3 27B}}}  
      & \textbf{Easy}   & \makecell{9.5{\scriptsize$\pm$4.1}} & \makecell{0.0{\scriptsize$\pm$0.0}} & \makecell{2.2{\scriptsize$\pm$3.9}} & \makecell{0.0{\scriptsize$\pm$0.0}} & \makecell{\textbf{11.8}{\scriptsize$\pm$0.0}} & \makecell{6.6{\scriptsize$\pm$2.1}} \\
      & \textbf{Medium}   & \makecell{0.0{\scriptsize$\pm$0.0}} & \makecell{0.0{\scriptsize$\pm$0.0}} & \makecell{4.2{\scriptsize$\pm$7.2}} & \makecell{0.0{\scriptsize$\pm$0.0}} & \makecell{0.0{\scriptsize$\pm$0.0}} & \makecell{1.3{\scriptsize$\pm$2.3}} \\
      & \textbf{Hard}   & \makecell{0.0{\scriptsize$\pm$0.0}} & \makecell{0.0{\scriptsize$\pm$0.0}} & \makecell{0.0{\scriptsize$\pm$0.0}} & \makecell{0.0{\scriptsize$\pm$0.0}} & \makecell{0.0{\scriptsize$\pm$0.0}} & \makecell{0.0{\scriptsize$\pm$0.0}} \\
      & \textbf{Total}   & \makecell{6.4{\scriptsize$\pm$2.8}} & \makecell{0.0{\scriptsize$\pm$0.0}} & \makecell{2.4{\scriptsize$\pm$2.1}} & \makecell{0.0{\scriptsize$\pm$0.0}} & \makecell{\textbf{8.0}{\scriptsize$\pm$0.0}} & \makecell{4.0{\scriptsize$\pm$1.0}} \\
    \bottomrule
  \end{tabular}
  \end{adjustbox}
\end{table*}


\textbf{MLLM Baselines}. We evaluated seven MLLMs on StarDojo-Lite task set: GPT-4.1 series~\cite{openai2025gpt4.1}(gpt-4.1-2025-04-14 \& gpt-4.1-mini-2025-04-14),  Claude 3.7 Sonnet~\cite{anthropic2025claude3.7} (claude-3-7-sonnet-20250219), Gemini 2.5 Pro~\cite{google2025gemini2.5} (gemini-2.5-pro-preview-03-25), from the closed-source community and Llama 4 Maverick~\cite{meta2025llama4} (Llama-4-Maverick-17B-128E-Instruct), Qwen2.5 VL~\cite{bai2025qwen2} (qwen2.5-vl-72b-instruct) and Gemma 3~\cite{team2025gemma} (gemma-3-27b-it) from the open-source community.

\textbf{Settings.} 
If not mentioned explicitly, all experiments are conducted under the following settings: Agents have access to both visual and textual observations for their decision-making process. Visual observations are provided at a resolution of 720P (1280×720). Textual observations include 7×7 agent-centered surrounding information and other global information like time, date and budget.
In addition to receiving observations from the current timestep, agents are also provided with history information from the previous timestep, enabling agents to reflect on past states and facilitating consistency in decision-making. Agents can output at most two skills as an action to be executed sequentially. After executing all the actions, the environment is paused until the agent outputs the next action. Each task is evaluated over three runs with a heuristic maximum of 30, 50 and 150 steps for easy, medium and hard level tasks. More details of settings are provided in Appendix~\ref{appendix:experiment_setting}

\subsection{Qualitative Analysis}

\textbf{Overall Results.}  As shown in Table~\ref{tab:agent_baseline}, a clear gap between flagship commercial models and open-source models can be observed. GPT-4.1, Gemini 2.5 Pro, and Claude 3.7 Sonnet all exceed a 10\% overall success rate, with GPT-4.1 achieving the highest at 12.7\%. In contrast, all open-source models remain below 8\%, with Llama 4 Maverick performing best, largely due to its larger model size.  Most successful completions are limited to easy tasks, whereas all models struggle significantly with medium and hard tasks, achieving near-zero success rates due to increased task complexity and longer sequences of required actions. Models show some proficiency in farming and crafting tasks but exhibit considerable difficulty in exploration, combat, and social interactions.

\textbf{Low-Level Control as Tool-Use.} All models demonstrate reasonable ability to control agents with non-trivial movements and interactions with actions clearly specified in the prompts. Larger models such as GPT-4.1, Gemini 2.5 Pro, Claude 3.7 Sonnet and Llama 4 Maverick perform significantly better than others on Crafting tasks, highlighting their superior tool-use capabilities and in-context understanding when handling more complex function calls. This contrast explains the weaker performance of smaller models like GPT-4.1 Mini and Gemma 3 27B, which often struggle to provide valid and accurate parameters when calling actions like \textit{choose\_option} and \textit{craft}.   However, all models continue to struggle in fast-paced tasks such as Combat, which demand dynamic control.

\textbf{Navigation is the Key Bottleneck.} We found that all models struggled most severely with navigation, which is the core skill required in the game. Due to limited image understanding, they consistently failed to accurately identify and locate target objects, building entrances or exits, and map transitions, which usually only appear in the provided image rather than the textual information. These shortcomings heavily impaired tasks that involve moving across scenarios or exploring areas beyond the immediate field of view. As a result, success rates were particularly low in Exploration and Social tasks, both of which often require cross-map navigation to locate targets. In contrast, agents are less influenced and perform relatively better in easy-level Farming tasks, since these are usually confined to a small farmland where target products remain within visual proximity and are directly accessible. 

\textbf{Long-Horizon Tasks Remain Far From Solved}. Medium and hard-level tasks can typically be decomposed into multiple simpler subtasks involving different category combinations. However, given the low success rates even on easy-level tasks, which serve as the atomic activities in the benchmark, all models remain far from being able to complete complex long-horizon tasks in StarDojo. This gap reveals substantial potential for more capable MLLM models.

\subsection{Error Analysis}
\begin{wrapfigure}{r}{0.4\linewidth}
    \centering
    \includegraphics[width=\linewidth]{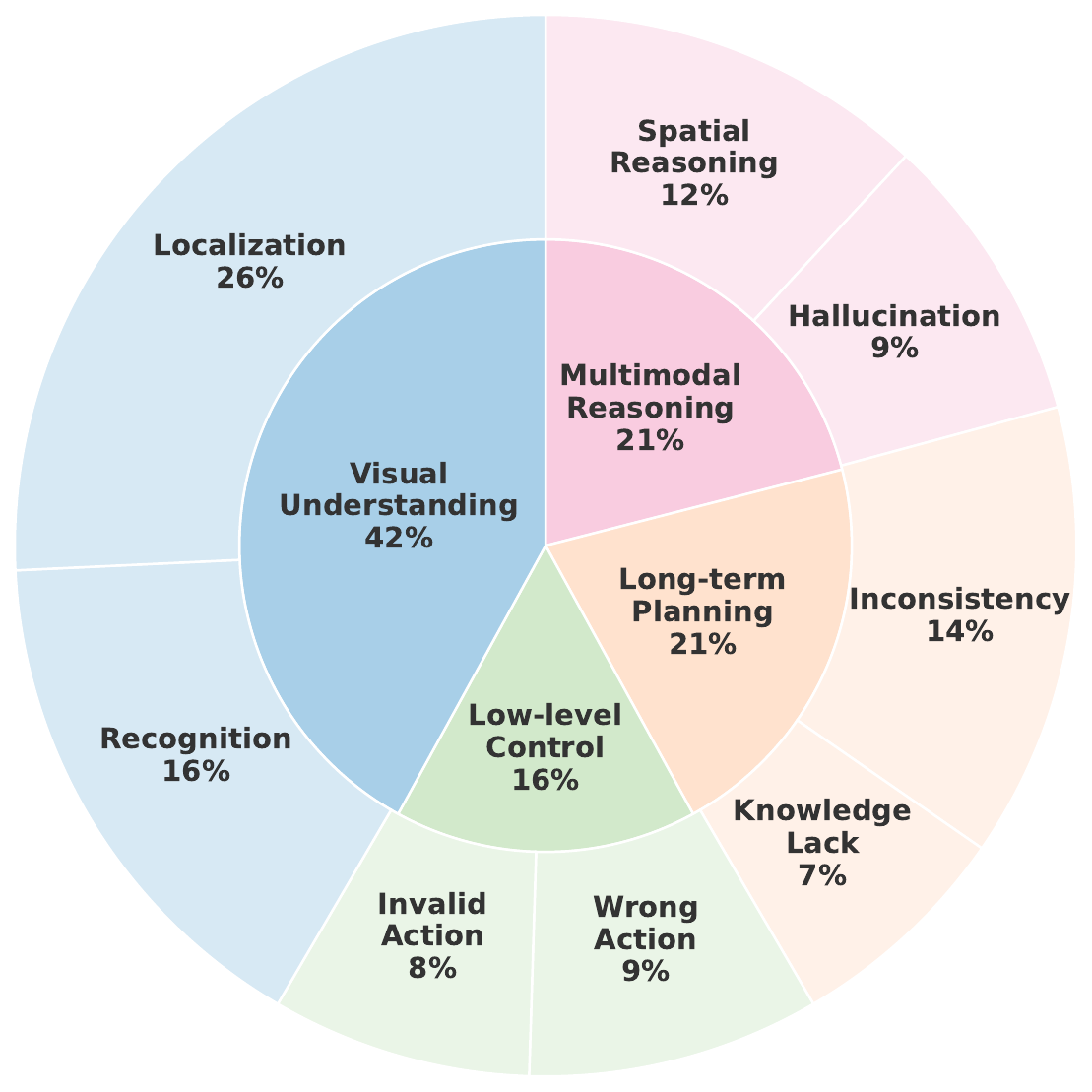}
    \caption{Error Analysis.}
    \label{fig:error_analysis}
\end{wrapfigure}
To further investigate models' behaviors, we conducted the error analysis on GPT-4.1-based agent to identify and categorize failure modes in StarDojo-Lite task set. For tasks with multiple errors, we report primary errors that block further progress.

The error analysis reveals several key failure modes, with the most significant being limited \textbf{Visual Understanding}, which accounts for 42\% of all errors. The model often struggles to reliably recognize target objects, many of which are only around 10x10 pixels shown in the image. And even when detected, it frequently fails to accurately determine their position relative to the character, hindering effective navigation and interaction. Additionally, models often struggle to interpret the status of tiles, even those immediately adjacent to the character, such as whether a tile is tilled, seeded, watered, or obstructed. The second major source of failure, at 21\%, is \textbf{Multimodal Reasoning}, where the model overly relies on its limited visual perception instead of integrating both visual and textual inputs, leading to confusion even when clear positional or status information is available in the text. Additionally, the model sometimes hallucinates task progress by incorrectly assuming the success of previous actions. \textbf{Long-Term Planning} issues also contribute to 21\% of failures, as the model, while capable of proposing reasonable subtasks, often abandons plans prematurely and switches strategies inconsistently, which undermines progress on longer-horizon tasks; this is sometimes compounded by insufficient domain knowledge, such as not knowing a crafting recipe or which NPC to interact with. Lastly, \textbf{Low-Level Control} accounts for a smaller but persistent 16\% of errors, where the model, despite generally choosing appropriate actions, occasionally selects incorrect or suboptimal ones, demonstrating instability in fine-grained execution, with occasional formatting or parameter errors. We provide more case studies and trajectory examples in Appendix~\ref{app:case_studies}.

\begin{figure*}[t]
    \centering
    \includegraphics[width=1\linewidth]{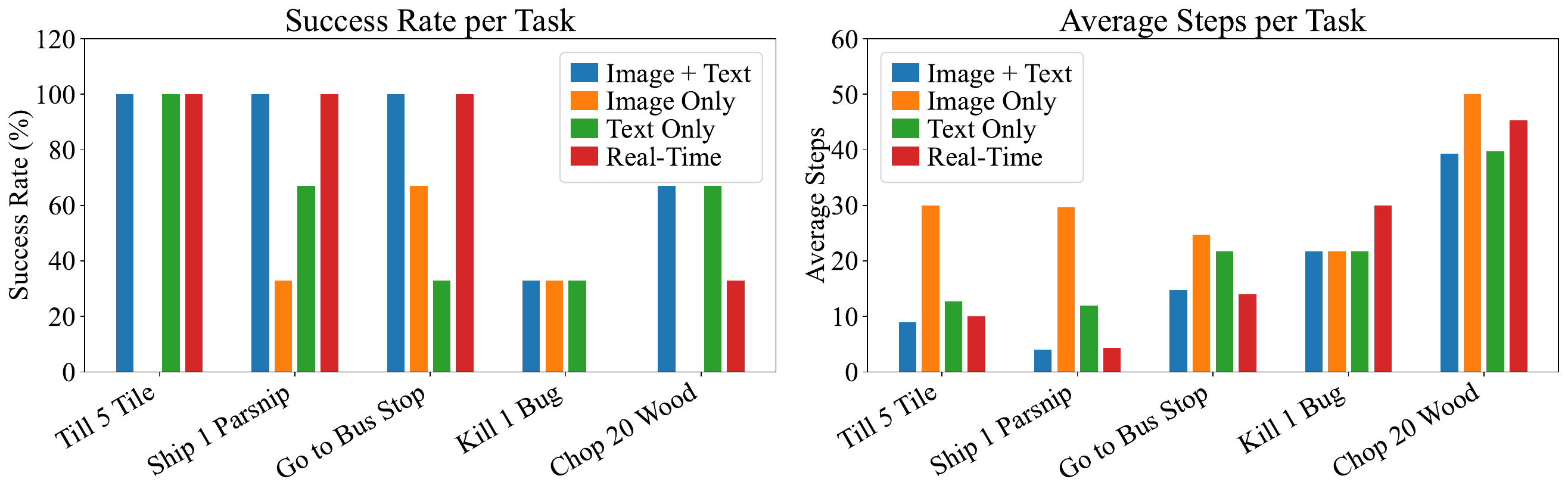}
    \caption{Ablation results of GPT-4.1-based agents on five representative tasks under four different settings: \textit{Image + Text} (default setting), \textit{Image Only}, \textit{Text Only}, and \textit{Real-Time} (without game pausing). Except for \textit{Chop 20 Wood}, which is a medium-level task with a maximum of 50 steps, the remaining four are easy-level tasks capped at 30 steps. Each task is evaluated over three runs.}
    \label{fig:ablation}
\end{figure*}

\subsection{Ablation Studies}
To demonstrate the flexible customization capabilities of StarDojo and enable more comprehensive evaluations of agent performance, we benchmark GPT-4.1 under the following three additional settings using five representative basic tasks.
1) \textbf{Image Only}: Agents receive only visual observations without textual state information, evaluating their capability to rely solely on visual cues.
2) \textbf{Text Only}: Agents receive textual state observations without visual input, restricting the agent's perception to local 7×7 grid-based information. This setting is particularly relevant for LLM agents.
3) \textbf{Real-time}: The environment progresses continuously during action generation, simulating real gameplay conditions where agents must plan and respond without game pausing.

As illustrated in Figure~\ref{fig:ablation}, removing textual input (\textit{Image Only}) significantly affects agents' performance across all tasks, reflecting the defect of base models' poor visual-based control, emphasizing textual information's importance in grounding detailed action decisions for the current stage of agents. 
On the other side, eliminating visual input (\textit{Text Only}) substantially reduces success in tasks that require navigation like \textit{Ship 1 Parsnip} and \textit{Go to Bus Stop}, demonstrating the essential contribution of visual cues to spatial reasoning and movement. Disabling the feature to pause the environment (\textit{Real-time}) remarkably affects performance across tasks demanding timely reactions or prolonged action sequences. 
For instance, in combat-related tasks such as \textit{Kill 1 Bug}, the bug keeps moving during the model's inference. It usually takes more than 10 seconds to get the GPT4.1's results through the API. The bug has already moved away after the model's inference and executed the actions based on the position 10 seconds ago. Similarly, in long-horizon tasks like \textit{Chop 20 Wood}, the in-game time keeps elapsed during models' inference, if passing, complete the task will usually take 2h in-game hours, if not pausing, it will take more than 12 in-game hours, which can extend into nighttime or span multiple in-game days that greatly delay and affect the completion of the task. Seldom previous benchmarks that emphasised the importance of evaluating the effect of real-timeness.  
These observations highlight the practical importance of real-time evaluation, a factor often overlooked in prior benchmarks. In many cases, agents may appear effective in offline evaluations but require substantially longer wall-clock time in real deployment settings. This issue is particularly pronounced for advanced agents built upon cutting-edge closed-source models accessed through APIs, where inference latency can become a bottleneck. In contrast, smaller open-source models typically offer faster inference but often suffer from substantially weaker task performance, leading to a fundamental trade-off between inference efficiency and capability.

\section{Limitations and Conclusion}
\label{sec:limitation_conclusion}
\textbf{Limitations.} This work has several potential limitations. 1) Although StarDojo is an open-source environment and benchmark, users need an official copy of Stardew Valley to run it. 2) Fishing, an optional gameplay, is currently not included in StarDojo, due to its nature as a complex, real-time mini-game. The operations are independent of other game controls, which is not applicable to evaluate MLLMs. 3) The benchmark primarily focuses on early- and mid-game content within the main valley map. Other advanced areas, such as the Desert and Ginger Island, are not yet supported. 4) Since “social” is a very broad concept, as the first benchmark that combines production and living, we do not intend to claim that StarDojo evaluates open-ended social intelligence across all aspects. Instead, it focuses on socially situated decision-making within an embodied environment. 5) Our evaluations were conducted mainly on StarDojo-Lite with seven models. A more comprehensive assessment across a broader range of tasks, models, and agentic approaches remains an important direction for future work.

\textbf{Conclusion.} We introduce StarDojo, a novel environment and benchmark designed to evaluate the open-ended behaviors of MLLM agents in Stardew Valley. StarDojo bridges the gap in existing environments by enabling comprehensive assessment of agents across both production and daily living activities within a simulated nature and society. Featuring a set of diverse tasks, StarDojo exposes significant challenges in current agents’ visual understanding, multimodal reasoning, long-term planning, and real-time inference, highlighting key areas for future research and development.


\section*{Acknowledgements}
This research is supported by the Ministry of Education, Singapore, under its Academic Research Fund Tier 1 (RG18/24). This research is also supported by the Singapore Ministry of Education (MOE) Academic Research Fund (AcRF) Tier 1 grant (Proposal ID: 23-SIS-SMU-037).

%
%
\bibliographystyle{splncs04}
\bibliography{main}

\clearpage

\appendix

\section{Introduction to Stardew Valley}
\label{appendix:intro_stardew}
Stardew Valley is an open-ended simulation RPG game where the player inherits a run-down farm and works to restore it. The game is presented in a 2D top-down perspective, offering a wider field of view compared to 3D games, which helps reduce operational difficulty and makes navigation smoother. The game does not have mandatory main storyline tasks and players can freely explore and live in the open-ended world.
\subsection{Realistic Gameplay Mechanism}
Stardew Valley is an open-ended simulation RPG game where the player inherits a run-down farm and works to restore it. The game is presented in a 2D top-down perspective, offering a wider field of view compared to 3D games, which helps reduce operational difficulty and makes navigation smoother. The game does not have mandatory main storyline tasks and players can freely explore and live in the open-ended world.

There are many mechanisms in Stardew Valley to simulate the real world, which raises additional challenges for players to plan reasonably according to these rules during the gameplay.

\textbf{Time and Daily Routine}. 
Each in-game day lasts about 14 minutes in real-time, running from 6 AM to 2 AM. Players must balance production, and social activities before exhaustion sets in. Staying up past 12 AM reduces energy the next day, and if players don’t sleep by 2 AM, they will pass out and lose gold or wake up at the clinic. Nightfall typically occurs between 6 and 8 PM, depending on the season, as the outdoors gradually darkens over time.

\textbf{Energy Management}. 
At the beginning, players have 270 energy points per day. Every action, from farming to mining, consumes energy. Energy can be restored by sleeping or eating food. Strategic time and energy management are essential to maximize productivity and also the main challenge in Stardew Valley.

\textbf{Seasons and Weather Effects}. 
The game also has four seasons
each lasting 28 days and affecting crop growth, fish availability, and town events. Weather varies daily, with rain saving time on watering crops, storms potentially damaging them, and snow limiting farming options. Some crops and activities are only available in specific seasons, requiring careful planning.

\subsection{Production}
Simulating real-world society, production activities are the main gameplay within the game, where players improve life quality through labor output.

\textbf{Labor Activities}. The game world is vast and diverse, featuring locations like Pelican Town, the Mines, the Beach and the forest, each offering unique activities such as mining, foraging, and fishing. Farming is central to gameplay, requiring players to plant, water, harvest crops and raise animals like cows and chickens for valuable products. Fishing and foraging provide additional resources, with seasonal variations and rare finds. Mining is crucial for gathering ores and materials, with progressively challenging levels and combat against monsters.

\textbf{Skill Progression and Crafting}. As players engage in these activities, they gradually improve their skills in Farming, Mining, Foraging, Fishing, and Combat. Gaining experience in each skill unlocks new crafting recipes, efficiency boosts, and profession choices that provide specialized benefits.
Crafting is an essential part of progression, allowing players to create tools, machines, and decorations that enhance farm efficiency and exploration. 
In addition to upgrading tools and structures, they can fully customize their farm and home, arranging decorations and personalizing interiors to create their ideal living space.

\subsection{Society}


Besides production, engaging with the community is a core aspect of the game.

\textbf{Festivals and Quests}.
Every season features unique festivals and events, such as the Egg Festival, Stardew Valley Fair, and Winter Star Festival. These events provide mini-games, rare items, and opportunities to strengthen relationships with villagers, adding depth to the community experience. Alongside festivals, quests play a crucial role in guiding players through different aspects of the game. 
NPCs also post daily requests on the town board, offering gold and friendship points for completing specific tasks.  By participating in festivals and completing quests, players engage more deeply with the world, fostering a sense of purpose and progression throughout their journey.

\textbf{Relationship and Friendship System}. 
The game also provides various NPCs with unique personalities, schedules, and heart events. Players can befriend them, give gifts, and even date or marry eligible characters. Higher friendship levels unlock new interactions, cutscenes, and benefits, such as helpful spouses assisting with farm chores. 


\textbf{Economy System}.
Stardew Valley features a comprehensive economic system where players generate income through various means, including farming, fishing, mining, and crafting. Players must manage their resources, reinvest in production, and make strategic decisions to ensure financial growth. The economy fluctuates based on seasonal demand, production choices, and market conditions. Strategic planning, investment in high-value goods, and efficient time management are essential to achieving long-term prosperity. The economic system provides depth and challenges players to optimize their approach to wealth generation and sustainability.



Stardew Valley serves as an ideal benchmark for decision-making agents in Production–Living Simulation. Its well-integrated systems of time management, resource allocation, economic planning, and social interaction provide a dynamic and complex environment that requires strategic thinking and adaptability. The game's structured yet open-ended nature makes it an excellent testbed for evaluating decision-making capabilities in simulated real-world conditions.

\section{StarDojo Environment}
\label{appendix:env}
Our environment is developed based on the game Stardew Valley, which offers comprehensive official mod development documentation. A large community of game enthusiasts has created and open-sourced their own mods. This provides significant convenience, as we can build a completely new mod on top of existing open-source mods to achieve the interaction between the LLM agent, RL agent, and the game.

Stardew Valley is available for purchase on Steam, and users must first buy and install the game to use our environment. Since our mod heavily relies on the official mod framework SMAPI, users will need to install both SMAPI and the mod we developed. This process can be easily carried out on Windows, macOS, and Linux systems with a graphical interface. Additionally, we have ensured compatibility for Linux systems without a graphical interface. After purchasing the game, users can install Stardew Valley through Steamcmd, the command-line tool of the Steam client to install and update dedicated servers for Steam games. We also provide installation commands for SMAPI, allowing users to install it directly. To use our developed mod, users simply need to copy it to the designated mod folder.
Since Stardew Valley is a graphical application, it cannot be directly opened on systems without a graphical interface. To address this limitation, we use the X virtual framebuffer (Xvfb), which supports all graphical operations in virtual memory without displaying any screen output. This directly ensures that our environment is compatible with various system and hardware configurations.

The interaction between our algorithm and Stardew Valley is achieved through the mod we developed, rather than simulating keyboard and mouse inputs. This approach allows us to open multiple game instances and control each one independently. Specifically, when launching multiple games, we assign a unique port number to each one. Actions provided by the algorithms being trained or tested are transmitted through this port to the corresponding game, and the returned information is received via the same port. Moreover, we use shared memory for each port to improve communication efficiency when necessary. By ensuring that the actions executed and information transmitted in all game processes do not interfere with one another, we enable parallel training and testing. To help researchers efficiently establish the initial state for each task, we provide the simulator APIs to configure the game environment.

\subsection{Observation Space}
\label{appendix:obs_space}
StarDojo provides a rich and flexible observation space to accommodate the diverse needs of various agents. The observation space includes both visual and textual observations, which can be customized and extended by developers to suit their specific goals.

\textbf{Visual Observation}. StarDojo leverages the game engine to directly retrieve rendered gameplay images, eliminating the need for inefficient screen-capture methods. This approach ensures high-fidelity visual observations that are consistent with the gameplay environment. The resolution of these images can be configured dynamically, ranging from 360P to 4K. Notably, the resolution scaling is not merely a resizing operation; as the image size increases, the field of view also expands, providing agents with a broader perspective of the game world. This feature is particularly useful for tasks requiring detailed spatial awareness or long-term planning.

\textbf{Textual Observation}. While visual observations are essential, recent work~\cite{tan2024cradle} highlights the challenges faced by state-of-the-art MLLMs in accurately interpreting the unique art style and precise manipulation requirements of Stardew Valley. To address this limitation, StarDojo provides structured textual observations through built-in APIs. These observations are designed to complement visual data, offering agents a more comprehensive understanding of the game state. The textual observations are organized into four main categories:

\begin{itemize}[leftmargin=*, topsep=0pt, partopsep=0pt, parsep=0pt, itemsep=0pt]
    \item \textbf{Character Information}: Includes health, energy, gold, position, inventory, location, facing direction, professions, skills (e.g., Farming, Mining, Combat, Fishing, Foraging), dating partners and spouse.
    \item \textbf{Surrounding Information}: Includes tile information within an N×N grid centered on the player’s position. Each tile in the grid contains detailed information about its contents and properties, such as debris, crops, NPCs, exits, buildings, furniture, terrain features, and other properties. The grid size can be adjusted to balance granularity and computational efficiency.
    \item \textbf{Map-level Global Information}: Provides global information within the current location map. Details such as crops, exits, NPCs, buildings, shop counters, furniture in rooms, and animals or pets on the farm are included as part of the map-level global information. By providing a structured overview of the entire map, this information enables the model to locate and reason about specific targets more efficiently, while reducing the need for exhaustive exploration of the environment.
    \item \textbf{Game-level Global Information}: Provides global information across the entire game. Game states such as time, day of month, season, year and weather are included in this part, along with other necessary information like the current menu information if any menu is showing.
\end{itemize}




As shown in our experiment, we selected a subset of the available information as the general configuration. Specifically, we adopted character information, surrounding information, and game-level global information to construct the observation space for Stardojo.
Although map-level global information can significantly boost agent performance in certain tasks, our goal is to evaluate the agent’s exploration ability based on visual input. Since map-level global information offers shortcuts for locating specific targets, it is deliberately excluded from our experimental setup.
This combination of visual and textual observations ensures that agents have access to both high-level contextual information and fine-grained environmental details, enabling more robust and informed decision-making.

\begin{figure}[htbp]
    \centering
    \includegraphics[width=0.45\textwidth]{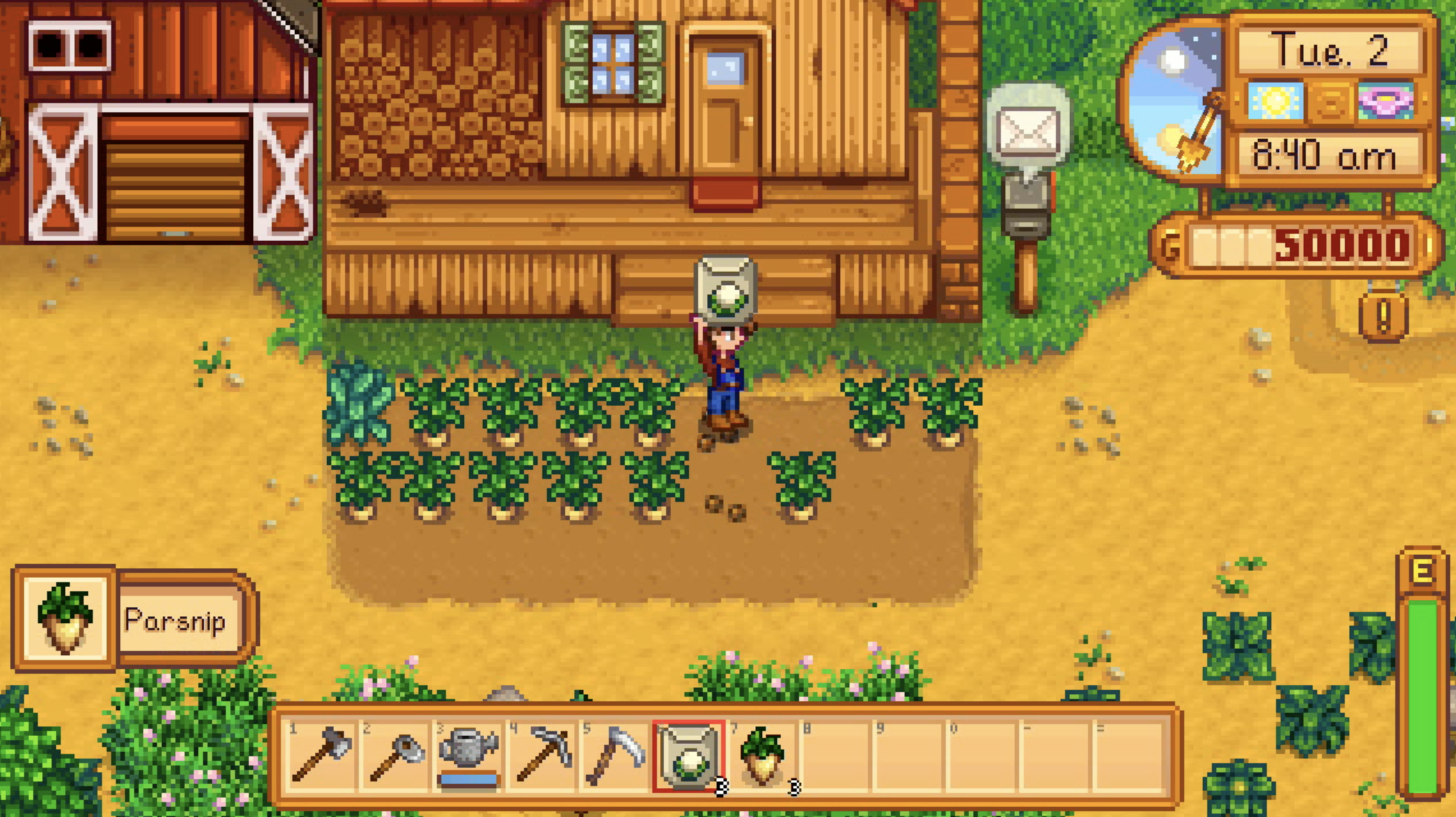} 
    \caption{Example of game screenshot as part of the observation space.}
    \label{fig:observation_example}
\end{figure}

The observation space captures a comprehensive snapshot of the game’s state in a JSON-like structure, with an additional screenshot RGB map. All relevant details are organized within the following nested objects. Figure \ref{fig:observation_example} shows an example of the screenshot included in the observation.

\begin{tcolorbox}[title=Observation Space Format, colback=white, colframe=black!75!gray, breakable]
\small
The observation space consists of the following structured fields:

\begin{itemize}[leftmargin=1.5em]
    \item \textbf{Health}: Integer representing the agent's current health.
    \item \textbf{Energy}: Float indicating the agent's current energy level.
    \item \textbf{Money}: Integer showing the amount of money the agent holds.
    \item \textbf{Current Time}: String formatted as \texttt{hh:mm AM/PM}.
    \item \textbf{Day}: Integer indicating the current day in the season.
    \item \textbf{Season}: String, one of \texttt{spring}, \texttt{summer}, \texttt{fall}, or \texttt{winter}.
    \item \textbf{Item in Your Hand}: A dictionary with fields:
        \begin{itemize}[noitemsep, topsep=0pt]
            \item \texttt{index}: Integer slot index.
            \item \texttt{currentitem}: String name of the item.
        \end{itemize}
    \item \textbf{Toolbar}: A list of 36 item slot descriptions in the format:
    
    \texttt{"slot\_index N: [Item Name] (quantity: Q)"} or \texttt{"slot\_index N: No item"}
    
    \item \textbf{Current Menu}: A dictionary with keys such as:
        \texttt{type}, \texttt{message}, \texttt{shopmenudata}, \texttt{animalsmenudata}, etc.
    \item \textbf{Surrounding Blocks}: A list of nearby tiles, each with:
        \begin{itemize}[noitemsep, topsep=0pt]
            \item \texttt{position}: A 2D integer coordinate offset relative to the agent.
            \item \texttt{object}: A list of string attributes (e.g., \texttt{Type: Dirt}, \texttt{Diggable: True}).
            \item (Optional) \texttt{npc on this tile}: Information about an NPC, if present.
        \end{itemize}
\end{itemize}
\end{tcolorbox}


\subsection{Action Space}
\label{appendix:action_space}
To better align with the actual action space of human players while simplifying redundant operations that do not contribute to the model’s decision-making capabilities, we designed a simplified minimal action space.

The action space defines the set of skills (or actions) that an agent can perform. Each action is implemented as a function with a specific call template and a thorough comment. The full list of actions is provided in \textbf{Table~\ref{tab:action_space_full}}, which details each available action along with its parameters and intended behavior. In our experiments, we excluded the \textit{navigate} action from the available action space. While \textit{navigate} provides a high-level shortcut for moving between maps, our primary objective is to evaluate the agent's realistic exploration abilities and its performance using an action space that more closely mirrors human interactions. 

\vspace{0.5em}  
\noindent
\renewcommand{\arraystretch}{1.2}
\small

\begin{table*}[htbp]
\caption{Complete Action Space with Call Templates and Parameter Descriptions}
\label{tab:action_space_full}
\centering
\begin{adjustbox}{max width=0.9\textwidth}
\begin{tabular}{@{}>{\raggedright\arraybackslash}p{4cm}@{\hspace{20pt}}p{\dimexpr\textwidth-4cm-6pt\relax}@{}}
\toprule
\textbf{Action} & \textbf{Description} \\
\midrule
\texttt{move(x, y)} & \textbf{Call Template:} \texttt{move(x = ..., y = ...)} \newline
\textbf{Parameters:} \texttt{x}, \texttt{y} – X and Y coordinates of the destination. \newline
Move to the position (x, y). \\
\midrule
\texttt{craft(item)} & \textbf{Call Template:} \texttt{craft(item = ...)} \newline
\textbf{Parameters:} \texttt{item} – The name of the item to craft. \newline
Craft an item based on its name. \\
\midrule
\texttt{use(direction)} & \textbf{Call Template:} \texttt{use(direction = ...)} \newline
\textbf{Parameters:} \texttt{direction} – A string: \texttt{up}, \texttt{right}, \texttt{down}, or \texttt{left}. \newline
Use an item in a specified direction. Requires proper positioning. \\
\midrule
\texttt{choose\_item(slot\_index)} & \textbf{Call Template:} \texttt{choose\_item(slot\_index = ...)} \newline
\textbf{Parameters:} \texttt{slot\_index} – Inventory index (0–35). \newline
Choose the item in the specified inventory slot. \\
\midrule
\texttt{interact(direction)} & \textbf{Call Template:} \texttt{interact(direction = ...)} \newline
\textbf{Parameters:} \texttt{direction} – A string: \texttt{up}, \texttt{right}, \texttt{down}, or \texttt{left}. \newline
Interact with an object or NPC in a specific direction. \\
\midrule
\texttt{choose\_option(option\_index, quantity, direction)} &
\textbf{Call Template:} \texttt{choose\_option(option\_index = ..., quantity = ..., direction = ...)} \newline
\textbf{Parameters:} \newline
\texttt{option\_index} – Index of the option to choose. \newline
\texttt{quantity} (optional) – Quantity of items to buy/sell. \newline
\texttt{direction} (optional) – \texttt{"in"} for buy/take, \texttt{"out"} for sell/put. \newline
Choose from a list of options, with optional quantity and direction. \\
\midrule
\texttt{attach\_item(slot\_index)} & \textbf{Call Template:} \texttt{attach\_item(slot\_index = ...)} \newline
\textbf{Parameters:} \texttt{slot\_index} – Inventory index (0–35). \newline
Attach the item in the given inventory slot. \\
\midrule
\texttt{unattach\_item()} & \textbf{Call Template:} \texttt{unattach\_item()} \newline
\textbf{Parameters:} None. \newline
Unattach the currently attached item. \\
\midrule
\texttt{menu(option, menu\_name)} & \textbf{Call Template:} \texttt{menu(option = ..., menu\_name = ...)} \newline
\textbf{Parameters:} \newline
\texttt{option} – \texttt{"open"} or \texttt{"close"} \newline
\texttt{menu\_name} – Menu name (e.g., \texttt{"map"}) \newline
Open or close a specific menu. \\
\midrule
\texttt{navigate(name)} & \textbf{Call Template:} \texttt{navigate(name = ...)} \newline
\textbf{Parameters:} \texttt{name} – Name of the location to navigate to. \newline
Navigate to a specified location. \\
\bottomrule
\end{tabular}
\end{adjustbox}
\end{table*}

\subsection{Task}
\label{appendix:task}

Our benchmark provides 1000 tasks organized into five categories:
\begin{itemize}[leftmargin=1.5em]
    \item \textbf{Farming}: Farming tasks can be broadly categorized into two types: cultivation (growing crops) and husbandry (raising animals). Easy tasks involve routine agricultural work such as clearing and tilling tiles, fertilizing, sowing seeds, watering plants, harvesting mature crops and animal products, as well as animal care—including feeding, grazing, and interaction. These simple, discrete operations test whether agents possess the most fundamental production capabilities. Medium tasks require agents to independently procure farming resources such as seeds, fertilizer, water, and hay. These resources can be obtained through foraging, crafting, or purchasing. Hard tasks typically span multiple days, requiring agents to perform daily routine farming activities. These activities form a cohesive production chain, following a specific sequence where each step is interdependent. For example, if a task involves growing a plant from seed to harvest, the agent must plant the seed in tilled dirt, water it daily over several days, and finally reap the crop. Any missed step, such as failing to water the plant on a given day, could delay or even prevent maturation. Thus, hard farming tasks demand that agents autonomously allocate time and resources efficiently, presenting significant tests of multi-step reasoning and long-term planning capabilities.

    \item \textbf{Crafting}: Crafting tasks encompass both fabrication and cooking. Most crafting activities simply require adequate raw materials, though some may need equipment like furnace or cookout kit. Typically, easy tasks provide all necessary materials and tools directly in the agent's inventory, requiring only proper execution of crafting procedures. Occasionally, agents might need to gather readily available resources like a few woods or stones. Medium tasks demand greater autonomy, which agents must identify, locate and acquire appropriate materials and equipment through careful planning. Hard tasks involve procuring diverse materials through demanding methods (like deep mining for ores), followed by complex, multi-stage processes like crafting intermediary components, testing an agent's comprehensive crafting capabilities.
  
    \item \textbf{Exploration}: Exploration tasks can be categorized into three types: map navigation/pathfinding, resource gathering in wilderness areas, and completing challenging in-game quests. Easy tasks may involve traveling to an accessible location or collecting specified items within a small nearby area. Medium tasks present greater complexity, some require venturing into more distant and hazardous environments (like the second floor of the mines), while others demand searching expansive zones (such as an entire forest) for randomly spawning resources. Hard tasks challenge agents to locate extremely rare resources with highly randomized spawn locations (like amethysts in mines). Additionally, built-in game quests that chain multiple sub-tasks of varying types, challenging enough to occasionally defeat even experienced human players also qualify as high-difficulty exploration objectives.

    \item \textbf{Combat}: Agents are tasked with eliminating a specified number of monsters in mines. As difficulty escalates, targets become both more formidable and numerous. Basic adversaries like slimes, bugs, and grubs present minimal threat, their predictable movement patterns, low health pools, and weak attacks make them easy to dispatch or evade. However, advanced creatures employ deadly specialties: flies attack with erratic, lightning-fast strikes; duggies ambush from subterranean positions; rock crabs retreat into impregnable armored stances. Defeating these requires dynamic positioning, tactical strike timing, and adaptive combat strategies.

    \item \textbf{Social}: Social tasks encompass two primary objectives: cultivating relationships with NPCs, and conducting transactional interactions with specialized NPCs (such as carpenter, blacksmith, etc.). Easy tasks require only basic interactions like conversations or gift-giving, and agents are teleported directly to designated transaction locations (e.g., at the counter of Pierre's General Store). Medium tasks demand agents autonomously select and navigate to appropriate venues. Hard tasks challenge agents to build high friendship levels with NPCs. This requires strategic planning to increase rapport through daily greetings, thoughtful gift-giving, and fulfilling requests. Each NPC possesses unique behavioral patterns and preferences. Agents must develop customized engagement strategies, as actions that delight one NPC (e.g., a favored gift) may offend another (e.g., a disliked item). This nuanced system tests agents' adaptive social intelligence.
\end{itemize}

Each category of tasks is encoded in a YAML file. This dictionary-like file format is both highly readable and convenient for processing by Python programs. The structure of a task is as follows:

\begin{tcolorbox}[title=Task Format, colback=white, colframe=black!75!gray, breakable]
\small
\textbf{sow\_5\_dirt\_with\_cauliflower\_seeds}: The key serves as both the name and the description of the task, and will be used as the prompt input to LLM.
\begin{itemize}[leftmargin=1.5em]
    \item \textbf{id}: A unique identifier for the task within its category.
    \item \textbf{object}: The target object of the task—such as item, location, character, or quest—that the player tries to acquire, reach, interact with, or complete in a specific quantity.
    \item \textbf{quantity}: The required number of the target object. For non-quantifiable objects, such as location and NPC, quantity is simply set to 1.
    \item \textbf{tool}: The tool required to complete the task.
    \item \textbf{save}: The initial game save for the task, which has pre-configured some common environmental settings to reduce frequent calls to simulator APIs.
    \item \textbf{init\_commands}: This is a list of commands to invoke the simulator APIs. After loading the task, StarDojo will automatically execute these commands one by one, invoking the corresponding API to fine-tune the initial conditions. Combined with the save file, StarDojo achieves efficient standardization for each task.
    \item \textbf{evaluator}: The type of evaluator to assess the task.
    \item \textbf{difficulty}: The difficulty level of the task.
\end{itemize}
\end{tcolorbox}


As shown in Table~\ref{tab:lite_tasks}, to facilitate efficient agent evaluation, we curate a representative smaller task suite, called StarDojo-Lite, comprising 100 core tasks from the full task collection, balancing coverage and practicality. The 100 tasks in StarDojo-Lite are listed in Table \ref{tab:task_specification}.

\begin{table*}[htbp]
\vspace{-10pt}
\caption{Complete list of tasks in StarDojo-Lite.}
\label{tab:task_specification}
\vspace{-10pt}
\centering
\small
\setlength{\tabcolsep}{0.5pt}
\begin{adjustbox}{max width=1\textwidth}
\begin{tabular}{@{}l|ccccccccc@{}}
\toprule
\textbf{Task} & \textbf{Category} & \textbf{ID} & \textbf{Object} & \textbf{Quantity} & \textbf{Tool} & \textbf{Save} & \textbf{Init Commands} & \textbf{Evaluator} & \textbf{Difficulty}\\
\midrule
clear\_10\_weeds\_with\_scythe & Farming & 0 & Weeds & 10 & Scythe & save\_new &  & clear & easy \\
clear\_5\_stone\_with\_pickaxe & Farming & 1 & Stone & 5 & Pickaxe & save\_new &  & clear & easy \\
clear\_30\_debris\_with\_scythe\_and\_pickaxe\_and\_axe & Farming & 2 & Debris & 30 & Scythe, Pickaxe, Axe & save\_new &  & clear & medium \\
till\_5\_tile\_with\_hoe & Farming & 3 & Tile & 5 & Hoe & save\_new &  & till & easy \\
fertilize\_5\_dirt\_with\_basic\_retaining\_soil & Farming & 4 & Dirt & 5 & Basic Retaining Soil & save\_farming & add\_item\_by\_name("Basic Retaining Soil", 5) & fertilize & easy \\
fertilize\_1\_dirt\_with\_speed\_gro & Farming & 5 & Dirt & 1 & Speed-Gro & save\_farming &  & fertilize & easy \\
sow\_5\_dirt\_with\_cauliflower\_seeds & Farming & 6 & Dirt & 5 & Cauliflower Seeds & save\_farming & add\_item\_by\_name("Cauliflower Seeds", 5) & sow & easy \\
sow\_1\_dirt\_with\_potato\_seeds & Farming & 7 & Dirt & 1 & Potato Seeds & save\_new & set\_time(time=900) & sow & medium \\
water\_5\_crop\_with\_watering\_can & Farming & 8 & Crop & 5 & Watering Can & save\_farming &  & water & easy \\
harvest\_5\_parsnip & Farming & 9 & Parsnip & 5 &  & save\_farming &  & harvest & easy \\
cultivate\_and\_harvest\_1\_garlic & Farming & 10 & Garlic & 1 &  & save\_new & add\_item\_by\_name("Garlic Seeds") & harvest & hard \\
pet\_3\_animal & Farming & 11 & Animal & 3 &  & save\_farming &  & pet & easy \\
pet\_8\_animal & Farming & 12 & Animal & 8 &  & save\_farming &  & pet & medium \\
open\_1\_deluxe\_coop & Farming & 13 & Deluxe Coop & 1 &  & save\_farming &  & open & easy \\
fill\_1\_pet\_bowl\_with\_watering\_can & Farming & 14 & Pet Bowl & 1 & Watering Can & save\_new &  & fill & easy \\
fill\_1\_feeding\_bench\_with\_hay & Farming & 15 & Feeding Bench & 1 & Hay & save\_farming & add\_item\_by\_name("Hay", 12) & fill & easy \\
harvest\_1\_egg & Farming & 16 & Egg & 1 &  & save\_farming &  & harvest & easy \\
harvest\_1\_milk\_with\_milk\_pail & Farming & 17 & Milk & 1 & Milk Pail & save\_farming & add\_item\_by\_name("Milk Pail") & harvest & easy \\
harvest\_3\_milk\_with\_milk\_pail & Farming & 18 & Milk & 3 & Milk Pail & save\_farming &  & harvest & hard \\
incubate\_1\_chicken\_with\_incubator & Farming & 19 & Chicken & 1 & Incubator & save\_farming &  & incubate & hard \\
earn\_50\_friendship\_with\_1\_cat & Farming & 20 & Cat & 50 &  & save\_new &  & friendship & hard \\
\midrule
craft\_1\_cherry\_bomb & Crafting & 0 & Cherry Bomb & 1 &  & save\_new & set\_time(time=900) & craft & medium \\
craft\_1\_wood\_fence & Crafting & 1 & Wood Fence & 1 &  & save\_new &  & craft & easy \\
craft\_1\_sprinkler & Crafting & 2 & Sprinkler & 1 &  & save\_new & \makecell{add\_item\_by\_name("Furnace"), add\_item\_by\_name("Copper Ore", 5),\\add\_item\_by\_name("Iron Ore", 5), add\_item\_by\_name("Coal", 2)} & craft & medium \\
craft\_1\_basic\_retaining\_soil & Crafting & 3 & Basic Retaining Soil & 1 &  & save\_new &  & craft & easy \\
craft\_1\_spring\_seeds & Crafting & 4 & Spring Seeds & 1 &  & save\_new &  & craft & hard \\
craft\_1\_field\_snack & Crafting & 5 & Field Snack & 1 &  & save\_new & \makecell{add\_item\_by\_name("Acorn"), add\_item\_by\_name("Maple Seed"),\\add\_item\_by\_name("Pine Cone")} & craft & easy \\
craft\_1\_torch & Crafting & 6 & Torch & 1 &  & save\_new &  & craft & easy \\
craft\_1\_furnace & Crafting & 7 & Furnace & 1 &  & save\_new & set\_time(time=900) & craft & medium \\
craft\_1\_chest & Crafting & 8 & Chest & 1 &  & save\_new & set\_time(time=900) & craft & medium \\
craft\_1\_scarecrow & Crafting & 9 & Scarecrow & 1 &  & save\_new & \makecell{add\_item\_by\_name("Wood", 50), add\_item\_by\_name("Coal"),\\add\_item\_by\_name("Fiber", 20)} & craft & easy \\
produce\_1\_copper\_bar\_with\_furnace & Crafting & 10 & Copper Bar & 1 & Furnace & save\_new & \makecell{add\_item\_by\_name("Furnace"), add\_item\_by\_name("Copper Ore", 5),\\add\_item\_by\_name("Coal")} & craft & easy \\
produce\_1\_refined\_quartz\_with\_furnace & Crafting & 11 & Refined Quartz & 1 & Furnace & save\_new &  & craft & hard \\
cook\_1\_fried\_egg\_with\_stove & Crafting & 12 & Fried Egg & 1 & Stove & save\_new & upgrade\_house(1), add\_item\_by\_name("Egg") & craft & easy \\
cook\_1\_salad\_with\_cookout\_kit & Crafting & 13 & Salad & 1 & Cookout Kit & save\_new &  & craft & hard \\
\midrule
go\_to\_bed & Exploration & 0 & Bed & 1 &  & save\_new &  & sleep & easy \\
go\_to\_coop & Exploration & 1 & Coop & 1 &  & save\_new &  & location & easy \\
go\_to\_bus\_stop & Exploration & 2 & BusStop & 1 &  & save\_new &  & location & easy \\
go\_to\_backwoods & Exploration & 3 & Backwoods & 1 &  & save\_new &  & location & easy \\
go\_to\_pierre's\_general\_store & Exploration & 4 & SeedShop & 1 &  & save\_new & set\_time(time=900) & location & easy \\
go\_to\_marnie's\_ranch & Exploration & 5 & AnimalShop & 1 &  & save\_new & set\_time(time=900) & location & easy \\
go\_to\_fish\_shop & Exploration & 6 & FishShop & 1 &  & save\_new & set\_time(time=900) & location & easy \\
go\_to\_carpenter's\_shop & Exploration & 7 & ScienceHouse & 1 &  & save\_new & set\_time(time=900) & location & easy \\
go\_to\_the\_mines\_2nd\_floor & Exploration & 8 & UndergroundMine2 & 1 &  & save\_new &  & location & medium \\
go\_to\_the\_mines\_5th\_floor\_by\_elevator & Exploration & 9 & UndergroundMine5 & 1 & Elevator & save\_new &  & location & medium \\
go\_to\_the\_mines\_10th\_floor & Exploration & 10 & UndergroundMine10 & 1 &  & save\_new &  & location & hard \\
chop\_10\_wood\_with\_axe & Exploration & 11 & Wood & 10 & Axe & save\_new &  & harvest & easy \\
chop\_20\_wood\_with\_axe & Exploration & 12 & Wood & 20 & Axe & save\_new &  & harvest & medium \\
forage\_1\_wild\_horseradish & Exploration & 13 & Wild Horseradish & 1 &  & save\_new & warp("forest") & harvest & medium \\
forage\_1\_daffodil & Exploration & 14 & Daffodil & 1 &  & save\_new & warp("town") & harvest & medium \\
forage\_1\_leek & Exploration & 15 & Leek & 1 &  & save\_new & warp("mountain") & harvest & medium \\
forage\_1\_clam & Exploration & 16 & Clam & 1 &  & save\_new & warp("beach") & harvest & easy \\
forage\_10\_hay\_with\_scythe & Exploration & 17 & Hay & 10 & Scythe & save\_new &  & silo & easy \\
forage\_1\_quartz & Exploration & 18 & Quartz & 1 &  & save\_new & warp\_mine(1) & harvest & medium \\
dig\_1\_cave\_carrot\_with\_hoe & Exploration & 19 & Cave Carrot & 1 & Hoe & save\_new & warp\_mine(13) & harvest & easy \\
mine\_1\_amethyst\_with\_pickaxe & Exploration & 20 & Amethyst & 1 & Pickaxe & save\_new & warp\_mine(1) & harvest & hard \\
mine\_1\_copper\_ore\_with\_pickaxe & Exploration & 21 & Copper Ore & 1 & Pickaxe & save\_new & warp\_mine(2) & harvest & easy \\
mine\_1\_coal\_with\_pickaxe & Exploration & 22 & Coal & 1 & Pickaxe & save\_new & warp\_mine(1) & harvest & medium \\
quit\_1\_quest & Exploration & 23 & Quest & 1 &  & save\_new &  & quit & easy \\
take\_1\_quest\_reward & Exploration & 24 & Quest Reward & 1 &  & save\_quests &  & reward & easy \\
complete\_1\_help\_wanted\_quest & Exploration & 25 & Help Wanted Quest & 1 &  & save\_new &  & complete\_help & hard \\
complete\_the\_story\_quest\_"introductions" & Exploration & 26 & 9 & 1 &  & save\_new &  & complete\_story & hard \\
complete\_the\_story\_quest\_"getting\_started" & Exploration & 27 & 6 & 1 &  & save\_quests &  & complete\_story & hard \\
\midrule
kill\_1\_green\_slime\_with\_rusty\_sword & Combat & 0 & Green Slime & 1 & Rusty Sword & save\_new & warp\_mine(2) & kill & easy \\
kill\_5\_green\_slime\_with\_rusty\_sword & Combat & 1 & Green Slime & 5 & Rusty Sword & save\_new & warp\_mine(2) & kill & medium \\
kill\_10\_green\_slime\_with\_rusty\_sword & Combat & 2 & Green Slime & 10 & Rusty Sword & save\_new & warp\_mine(2) & kill & hard \\
kill\_1\_bug\_with\_rusty\_sword & Combat & 3 & Bug & 1 & Rusty Sword & save\_new & warp\_mine(2) & kill & easy \\
kill\_5\_bug\_with\_rusty\_sword & Combat & 4 & Bug & 5 & Rusty Sword & save\_new & warp\_mine(2) & kill & medium \\
kill\_10\_bug\_with\_rusty\_sword & Combat & 5 & Bug & 10 & Rusty Sword & save\_new & warp\_mine(2) & kill & hard \\
kill\_1\_fly\_with\_rusty\_sword & Combat & 6 & Fly & 1 & Rusty Sword & save\_new & warp\_mine(2) & kill & medium \\
kill\_1\_duggy\_with\_rusty\_sword & Combat & 7 & Duggy & 1 & Rusty Sword & save\_new & warp\_mine(6) & kill & medium \\
kill\_1\_grub\_with\_rusty\_sword & Combat & 8 & Grub & 1 & Rusty Sword & save\_new & warp\_mine(15) & kill & easy \\
kill\_5\_grub\_with\_rusty\_sword & Combat & 9 & Grub & 5 & Rusty Sword & save\_new & warp\_mine(15) & kill & medium \\
kill\_10\_grub\_with\_rusty\_sword & Combat & 10 & Grub & 10 & Rusty Sword & save\_new & warp\_mine(15) & kill & hard \\
kill\_1\_rock\_crab\_with\_rusty\_sword & Combat & 11 & Rock Crab & 1 & Rusty Sword & save\_new & warp\_mine(2) & kill & medium \\
\midrule
ship\_1\_parsnip\_with\_shipping\_bin & Social & 0 & Parsnip & 1 & Shipping Bin & save\_new & add\_item\_by\_name("Parsnip") & sell & easy \\
purchase\_5\_beer & Social & 1 & Beer & 5 &  & save\_new & warp\_shop("gus") & purchase & easy \\
purchase\_1\_muscle\_remedy & Social & 2 & Muscle Remedy & 1 &  & save\_new & set\_time(time=900) & purchase & medium \\
sell\_5\_parsnip\_to\_pierre & Social & 3 & Parsnip & 5 &  & save\_new & warp\_shop("pierre"), add\_item\_by\_name("Parsnip", 5) & sell & easy \\
sell\_1\_parsnip\_to\_pierre & Social & 4 & Parsnip & 1 &  & save\_new & add\_item\_by\_name("Parsnip"), set\_time(time=900) & sell & medium \\
upgrade\_to\_copper\_pickaxe & Social & 5 & Copper Pickaxe & 1 &  & save\_new & add\_item\_by\_name("Copper Bar", 5) & upgrade\_tool & hard \\
break\_5\_geode & Social & 6 & Geode & 5 &  & save\_new & warp\_shop("clint"), add\_item\_by\_name("Geode", 5) & break & easy \\
purchase\_joja\_membership & Social & 7 & Joja Membership & 1 &  & save\_new & warp("joja", 21, 26) & jojamart & easy \\
purchase\_minecarts\_development\_project & Social & 8 & Minecarts Repaired & 1 &  & save\_new & joja\_membership(), set\_time(time=900) & jojamart & medium \\
upgrade\_to\_large\_pack & Social & 9 & Large Pack & 1 &  & save\_new & warp\_shop("pierre") & backpack & easy \\
purchase\_1\_chicken & Social & 10 & Chicken & 1 &  & save\_new & warp\_shop("marnie") & purchase\_animal & easy \\
sell\_1\_chicken & Social & 11 & Chicken & 1 &  & save\_new &  & sell\_animal & easy \\
build\_1\_big\_coop & Social & 12 & Big Coop & 1 &  & save\_new & \makecell{warp\_shop("robin"), add\_item\_by\_name("Wood", 400),\\add\_item\_by\_name("Stone", 150)} & build & easy \\
move\_1\_coop & Social & 13 & Coop & 1 &  & save\_new & set\_time(time=900) & move & medium \\
upgrade\_farmhouse & Social & 14 & Farmhouse & 1 &  & save\_new & warp\_shop("robin"), add\_item\_by\_name("Wood", 450) & upgrade\_farmhouse & easy \\
demolish\_1\_shipping\_bin & Social & 15 & Shipping Bin & 1 &  & save\_new & set\_time(time=900) & demolish & medium \\
talk\_to\_alex & Social & 16 & Alex & 1 &  & save\_new & set\_time(time=800) & talk & easy \\
talk\_to\_sebastian & Social & 17 & Sebastian & 1 &  & save\_new & set\_time(time=1500) & talk & easy \\
talk\_to\_vincent & Social & 18 & Vincent & 1 &  & save\_new & set\_time(time=900) & talk & easy \\
give\_abigail\_1\_amethyst & Social & 19 & Abigail & 1 & Amethyst & save\_new & add\_item\_by\_name("Amethyst", 1, 4), set\_time(time=900) & gift & easy \\
give\_haley\_1\_coconut & Social & 20 & Haley & 1 & Coconut & save\_new & add\_item\_by\_name("Coconut", 1, 4), set\_time(time=1100) & gift & easy \\
give\_jas\_1\_fairy\_rose & Social & 21 & Jas & 1 & Fairy Rose & save\_new & add\_item\_by\_name("Fairy Rose", 1, 4), set\_time(time=900) & gift & easy \\
give\_jodi\_1\_pancakes & Social & 22 & Jodi & 1 & Pancakes & save\_new & add\_item\_by\_name("Pancakes", 1, 4), set\_time(time=900) & gift & easy \\
earn\_100\_friendship\_with\_elliott & Social & 23 & Elliott & 100 &  & save\_new & add\_item\_by\_name("Pomegranate", 1, 4), set\_time(time=1100) & friendship & medium \\
earn\_200\_friendship\_with\_harvey & Social & 24 & Harvey & 200 &  & save\_new &  & friendship & hard \\
\bottomrule
\end{tabular}
\end{adjustbox}
\end{table*}

\clearpage

\subsection{Simulator APIs}
To establish the initial state for each task, we develop a set of simulator APIs that allow fine-grained customization of the game environment. Many tasks require strict initial conditions and world settings, such as resources (e.g., sufficient crop seeds in inventory), weather conditions (e.g., a rainy day), and game progress (e.g., unlocking the mines). The simulator APIs can configure these task-specific conditions, endowing each task with a unique environment setup. This mechanism significantly expands the benchmark’s diversity and flexibility, allowing varied and nuanced task designs that closely mirror the dynamic production-living settings in human society.

To efficiently standardize our tasks, we also create a series of tailored game save files, which have pre-configured some common environmental settings using simulator APIs. These files reduce the excessively frequent calls to simulator APIs in real time, improving the efficiency of task execution. At the beginning of each task, StarDojo automatically loads the corresponding saved game and invokes specific simulator APIs, ensuring all necessary prerequisites are appropriately configured.

The complete APIs are as follows:

\begin{tcolorbox}[title=Simulator APIs, colback=white, colframe=black!75!gray, breakable]
\small
\begin{itemize}[leftmargin=1.5em]
    \item \textbf{Player Settings}
        \begin{itemize}[leftmargin=1.5em]
            \item \textbf{set\_base\_health(amount: int)}: Set the health capacity of the player.
            \item \textbf{set\_health(amount: int)}: Set the current health of the player.
            \item \textbf{set\_base\_energy(amount: int)}: Set the energy capacity of the player.
            \item \textbf{set\_energy(amount: int)}: Set the current energy of the player.
            \item \textbf{set\_inventory\_size(size: int)}: Set the inventory size.
            \item \textbf{clear\_inventory()}: Clear the inventory.
            \item \textbf{set\_money(amount: int)}: Set the amount of money that the player possesses.
            \item \textbf{add\_item\_by\_id(id: str, count: int, quality: int)}: Add the specific item of given count and quality (e.g., 0, 1) to inventory by ID.
            \item \textbf{add\_item\_by\_name(name: str, count: int, quality: int)}: Add the specific item of given count and quality to inventory by name.
            \item \textbf{lookup(name: str)}: Look up and print the item ID by name.
            \item \textbf{current\_position()}: Print the current position of the player.
            \item \textbf{add\_recipe(type: str, recipe: str)}: Teach the player the specific crafting / cooking recipe.
            \item \textbf{set\_max\_luck()}: Set player's luck to the maximum.
            \item \textbf{print\_luck()}: Print player's luck.
        \end{itemize}
    \item \textbf{Surrounding Settings}
        \begin{itemize}[leftmargin=1.5em]
            \item \textbf{world\_clear(entity: str, location: str)}: Remove all entities of the given type (e.g., "crops", "trees") from a location.
            \item \textbf{set\_terrain(terrain: str, id: str, x: int, y: int)}: Set the terrain feature of the given tile.
            \item \textbf{place\_item(item: str, type: str, x: int, y: int)}: Place the specific item on the given tile.
            \item \textbf{remove\_item(x: int, y: int)}: Remove the item on the given tile.
            \item \textbf{place\_crop(crop: str, x: int, y: int)}: Place the specific crop on the given tile.
            \item \textbf{grow\_crop(day: int, x: int, y: int)}: Grow the crop on the given tile for a specific number of days.
            \item \textbf{grow\_tree(day: int, x: int, y: int)}: Grow the tree on the given tile for a specific number of days.
            \item \textbf{build(type: str, force: bool, x: int, y: int)}: Build the specific building at the given coordinate.
            \item \textbf{build\_stable(x: int, y: int)}: Build a stable at the given coordinate.
            \item \textbf{move\_building(x\_source: int, y\_source: int, x\_dest: int, y\_dest: int)}: Move the building from the source coordinate to the destination coordinate.
            \item \textbf{remove\_building(x: int, y: int)}: Remove the building at the given coordinate.
            \item \textbf{upgrade\_house(level: int)}: Upgrade the farmhouse to the given level.
        \end{itemize}
    \item \textbf{Character Settings}
        \begin{itemize}[leftmargin=1.5em]
            \item \textbf{spawn\_pet(type: str, breed: str, name: str, x: int, y: int)}: Spawn a pet of given type (e.g., "cat", "dog"), breed (e.g., "0", "1"), and name on a tile.
            \item \textbf{spawn\_animal(type: str, name: str)}: Spawn an animal of given type and name in the animal house.
            \item \textbf{grow\_animal(name: str)}: Set the specific animal in current location to day 1 of adulthood, unless already adult.
            \item \textbf{animal\_friendship(name: str, friendship: int)}: Set the friendship with the specific animal.
            \item \textbf{npc\_friendship(npc: str, friendship: int)}: Set the friendship with the specific NPC.
            \item \textbf{all\_npc\_friendship(friendship: int)}: Set the friendship with all NPCs.
            \item \textbf{dating(npc: str)}: Make the specific NPC be the player's boyfriend / girlfriend.
        \end{itemize}
    \item \textbf{Location Settings}
        \begin{itemize}[leftmargin=1.5em]
            \item \textbf{warp(location: str, x: int, y: int)}: Warp the player to given location and position.
            \item \textbf{warp\_mine(level: int)}: Warp the player to given mine level.
            \item \textbf{warp\_volcano(level: int)}: Warp the player to given volcano level.
            \item \textbf{warp\_home()}: Warp the player back home.
            \item \textbf{warp\_shop(npc: str)}: Warp the player to the shop run by the given NPC.
            \item \textbf{warp\_character(npc: str, location: str, x: int, y: int)}: Warp the specific NPC to given location and position.
        \end{itemize}
    \item \textbf{World Settings}
        \begin{itemize}[leftmargin=1.5em]
            \item \textbf{set\_date(year: int, season: str, day: int)}: Set the date.
            \item \textbf{set\_time(time: int)}: Set the current time.
            \item \textbf{rain()}: Set the weather to rainy.
        \end{itemize}
    \item \textbf{Progression Settings}
        \begin{itemize}[leftmargin=1.5em]
            \item \textbf{set\_deepest\_mine\_level(level: int)}: Set the deepest mine level reached by the player.
            \item \textbf{set\_monster\_stats(monster: str, kills: int}: Set the kill stats for the specific monster to the given value.
            \item \textbf{print\_monster\_stats(monster: str)}: Print the kill stats for the specific monster.
            \item \textbf{start\_quest(id: str)}: Start the quest of given ID.
            \item \textbf{start\_help\_quest(type: str)}: Start a random help wanted quest of given type.
            \item \textbf{complete\_quest(id: str)}: Complete the quest of given ID.
            \item \textbf{joja\_membership()}: Give the player JojaMart membership.
            \item \textbf{spawn\_junimo\_note(id: str)}: Spawn the junimo note of given ID in the Community Center.
            \item \textbf{mark\_bundle(id: str)}: Mark the completion of the specific bundle.
            \item \textbf{complete\_room\_bundles(id: str)}: Complete all bundles in the specific room of the Community Center.
            \item \textbf{community\_development(id: str)}: Complete the community development project of given ID.
            \item \textbf{receive\_mail(mail: str)}: Add the specific mail to mailbox.
            \item \textbf{trigger\_event(id: str)}: Trigger the event of given ID.
            \item \textbf{seen\_event(id: str, see\_or\_forget: bool)}: Mark the viewed / unviewed status of the specific event.
            \item \textbf{load\_save(save: str)}: Load the game save of given name.
        \end{itemize}
\end{itemize}
\end{tcolorbox}

\subsection{Evaluation}
Given the scale of our extensive task set, it is imperative to design an efficient and reusable evaluation mechanism that not only accurately monitors task progression but also delivers immediate reward feedback to agents. Therefore, we implement a unified evaluation system based on textual observation comparison. The evaluation workflow follows a consistent pattern across all tasks, as outlined in the following steps and Algorithm \ref{algorithm}:
\begin{itemize}[leftmargin=1.5em]
    \item \textbf{Maintain the previous observation}: Store the agent's prior textual observation to enable temporal comparison.
    \item \textbf{Capture the current observation}: Acquire the latest textual observation.
    \item \textbf{Compare two observations}: Based on the task's evaluator type (e.g., harvest, sell) and target object (e.g., item, NPC), the system detects related game state changes, such as items in the inventory and surrounding tiles, to quantify task progress.
    \item \textbf{Accumulate incremental progress:} Accumulate incremental changes captured per step into a sum.
    \item \textbf{Check completion criterion}: Validate whether predefined success conditions (e.g., quantity thresholds, event triggers) are met.
\end{itemize}

\begin{algorithm}
\caption{Task Evaluation}
\begin{algorithmic}[1]
\Function{Evaluate}{$obs$}
  \If{$last\_obs$ is null}
    \State $last\_obs \gets obs$
    \State \Return $\{\text{completed}: \text{False},\; \text{current\_quantity}: 0\}$
  \EndIf
  \State $quantity\_change \gets \Call{Compare}{evaluator, object, obs, last\_obs}$
  \State $last\_obs \gets obs$
  \State $current\_quantity \gets current\_quantity + quantity\_change$
  \If{$current\_quantity \geq quantity$}
    \State $completed \gets \text{True}$
  \Else
    \State $completed \gets \text{False}$
  \EndIf
  \State \Return $\{\text{completed}: completed,\; \text{current\_quantity}: current\_quantity\}$
\EndFunction
\end{algorithmic}
\label{algorithm}
\end{algorithm}

The evaluation system relies on the comprehensive observation space, allowing for step-by-step tracking of progress toward the task goal. This incremental inspection approach avoids differences caused by varying initial conditions, making the proposed evaluation mechanism more generalizable. The evaluation output provides standardized metrics, including completion status and current progress. These metrics also serve as reward feedback and, along with observations, are passed to the agents to prompt them to adjust their behavior.

\section{Experiment Settings}
\label{appendix:experiment_setting}
If not mentioned explicitly, all experiments are conducted under the following settings: Agents have access to both visual and textual observations for their decision-making process. Visual observations are provided at a resolution of 720p (1280×720 pixels). Textual observations include detailed information about current state. Specifically, there are information about the player itself, including health, energy, gold, chosen item, and inventory information. There are also global information, which are current time, current day, season, and current open menu. Finally, we provide 7×7 agent-centered surrounding information, containing details about each tile, ranging from the terrain, debris, buildings, object, exits, NPCs, and furniture information, to other tile properties encoded in the game. In addition to receiving observations from the current timestep, agents are also provided with action and visual information from the previous timestep. Incorporating previous timestep information enables agents to reflect on past states and facilitate consistency in decision-making. All the agents can output at most two skills as an action to be executed sequentially. After executing all the actions, the environment remains paused until the agent outputs the next action. All experiments are repeated three times to ensure reliability.

\begin{tcolorbox}[title=Prompt Used in Main Experiment,colback=white,colframe=black!75!gray,breakable]
\small
You are a helpful AI assistant integrated with 'Stardew Valley' on the PC, equipped to handle various tasks in the game. Your advanced capabilities enable you to process and interpret gameplay screenshots and other relevant information. By analyzing these inputs, you gain a comprehensive understanding of the current context and situation within the game. Utilizing this insight, you are tasked with identifying the most suitable in-game action to take next, given the current task. You control the game character and can execute actions from the available action set. Upon evaluating the provided information, your role is to articulate the precise action you would deploy, considering the game's present circumstances, and specify any necessary parameters for implementing that action.

Here is some helpful information to help you make the decision.

Your Current task is: <\$task\_description\$> \\

Basic knowledge about the environment: \\
\begin{enumerate}
\item Hoe is used to till the soil, Watering Can is used to water the soil, Pickaxe is used to break rocks, Axe is used to chop trees, Scythe is used to harvest crops.
\item When you want to go through a door, move in front of it by 1 tile, and interact towards it.
\item Please go to bed at night (after 18:00) even if your task is not yet complete!
\item Call interact(direction) with a box, a shipping bin or anything else. Call use(direction) to use an item or tool in your inventory.
\end{enumerate}

Health: <\$health\$> \\

Energy: <\$energy\$> \\

Money: <\$money\$> \\

Current Time: <\$time\$> \\

Day: <\$day\$> \\

Season: <\$season\$> \\

Item in your hand: <\$chosen\_item\$> \\

Toolbar of items which you can choose from: <\$toolbar\_information\$> \\

Current menu: <\$current\_menu\$> \\

Surrounding blocks (Objects surrounds you): <\$surroundings\$> \\

Valid action set in Python format to select the next action: <\$skill\_library\$> \\

Last executed action: <\$pre\_action\$> \\

<\$image\_introduction\$> \\

Based on the above information, analyze the current situation and provide the reasoning for what you should do for the next step to complete the task. Then, you should output the exact action you want to execute in the game. You should respond to me with: \\

Reasoning: You should think step by step and provide detailed reasoning to determine the next action executed on the current state of the task. You need to answer the following questions step by step. You cannot miss the last question:
\begin{enumerate}
    \item Is there an open menu? What is the current menu saying? What are the options? Which one should you choose or should you exit the menu? Use choose\_option to make a choice.
    \item What is the current map? Do you need to move to another map?
    \item Refer to surroundings, what are the important tiles? What are their positions?
    \item You are always at (0, 0). You can only affect points adjacent to you, such as (0,1), (0,-1), (1,0), (-1,0). Is your target at these positions? If not, move next to it first.
    \item Analyze the information in the toolbar. Does it contain all the necessary items for completing the task? What is the current item?
    \item When calling use or interact, you need to decide the direction. For example, if you are at (x,y): \\
    \quad - call interact("up") or use("up") to interact with or use against (0,-1) \\
    \quad - call interact("right") or use("right") to interact with or use against (1,0) \\
    \quad - call interact("down") or use("down") to interact with or use against (0,1) \\
    \quad - call interact("left") or use("left") to interact with or use against (-1,0)
    \item What is the current image showing? What additional information can be learned from the image?
    \item Do all the selected actions exist in the valid action set? If no, regenerate the actions and give the reasons.
\end{enumerate}

Actions: The requirements that the generated action needs to follow. The best action, or short sequence of actions without gaps, to execute next to progress in achieving the goal. Pay attention to the names of the available skills and to the previous skills already executed, if any. You should also pay more attention to the following action rules:
\begin{enumerate}
    \item You should output actions in Python code format and specify any necessary parameters to execute that action. If the function has parameters, you should also include their names and decide their values, like move(x=0, y=1). If it does not have a parameter, just output the action, like unattch\_item().
    \item You can only output at most 2 actions in the output.
    \item If you want to interact with the objects in the toolbar, you need to make sure that the target object is already selected. You need to use choose\_item() to select them before executing use().
    \item If you want to plant a seed or harvest a mature crop, please use interact(). If you want to use tools, like axe, hoe, watering can, pickaxe and scythe, please use use().
    \item Your action should strictly follow the analysis in the reasoning. Do not output any additional action not mentioned in the reasoning.
\end{enumerate}

You should only respond in the format described below, and you should not output comments or other information. \\
Reasoning: \\
1. ... \\
2. ... \\
3. ... \\
Actions:\\
\begin{verbatim}
```
python
    action(args1=x,args2=y)
```
\end{verbatim}
\end{tcolorbox}

\textbf{Conclusion}

The above cases collectively reveal key limitations of current large language models when deployed in grounded, spatially structured environments like Stardew Valley. Despite having access to both visual and textual information, the model consistently exhibits hallucinations in spatial understanding — including confusion about direction, misjudgment of proximity, and failure to plan indirect paths around obstacles.

These behaviors suggest that LLMs, while capable in language-based reasoning, still lack robust internal representations of space and geometry. Moreover, the persistence of such errors even with explicit instructions points to fundamental weaknesses in grounding language to actionable physical reasoning. Bridging this gap will require future work in multimodal integration, spatial memory, and instruction-following mechanisms tailored to embodied agents. Our findings underscore the importance of evaluating LLMs not just on language tasks, but within interactive environments where spatial and physical reasoning are essential.

\section{Financial Cost}
The financial costs of evaluating different closed-source MLLMs on StarDojo-Lite for one round are summarized in Table \ref{tab:cost}.
\begin{table}[t]
    \caption{Financial costs of running one round on StarDojo-Lite with different closed-source MLLMs.}
    \centering
    \small
    \setlength{\tabcolsep}{8pt}
    \begin{adjustbox}{width=\linewidth}
    \begin{tabular}{@{}l|lr@{}}
    \toprule
    \textbf{MLLM} & \textbf{Model Name} & \textbf{Cost}\\
    \midrule
    GPT-4.1 & gpt-4.1-2025-04-14 & \$20.6 \\
    Gemini 2.5 Pro & gemini-2.5-pro-preview-03-25 & \$25.8 \\
    Claude 3.7 Sonnet & claude-3-7-sonnet-20250219 & \$38.7 \\
    GPT-4.1 mini & gpt-4.1-mini-2025-04-14 & \$4.3 \\
    \bottomrule
    \end{tabular}
    \end{adjustbox}
    \label{tab:cost}
\end{table}

\clearpage

\section{Case Studies}
\label{app:case_studies}



In this section, we provide additional visualization cases of agent performance in StarDojo to better illustrate the behaviors and limitations of current models.

\begin{figure}
    \centering
    \includegraphics[width=1\linewidth]{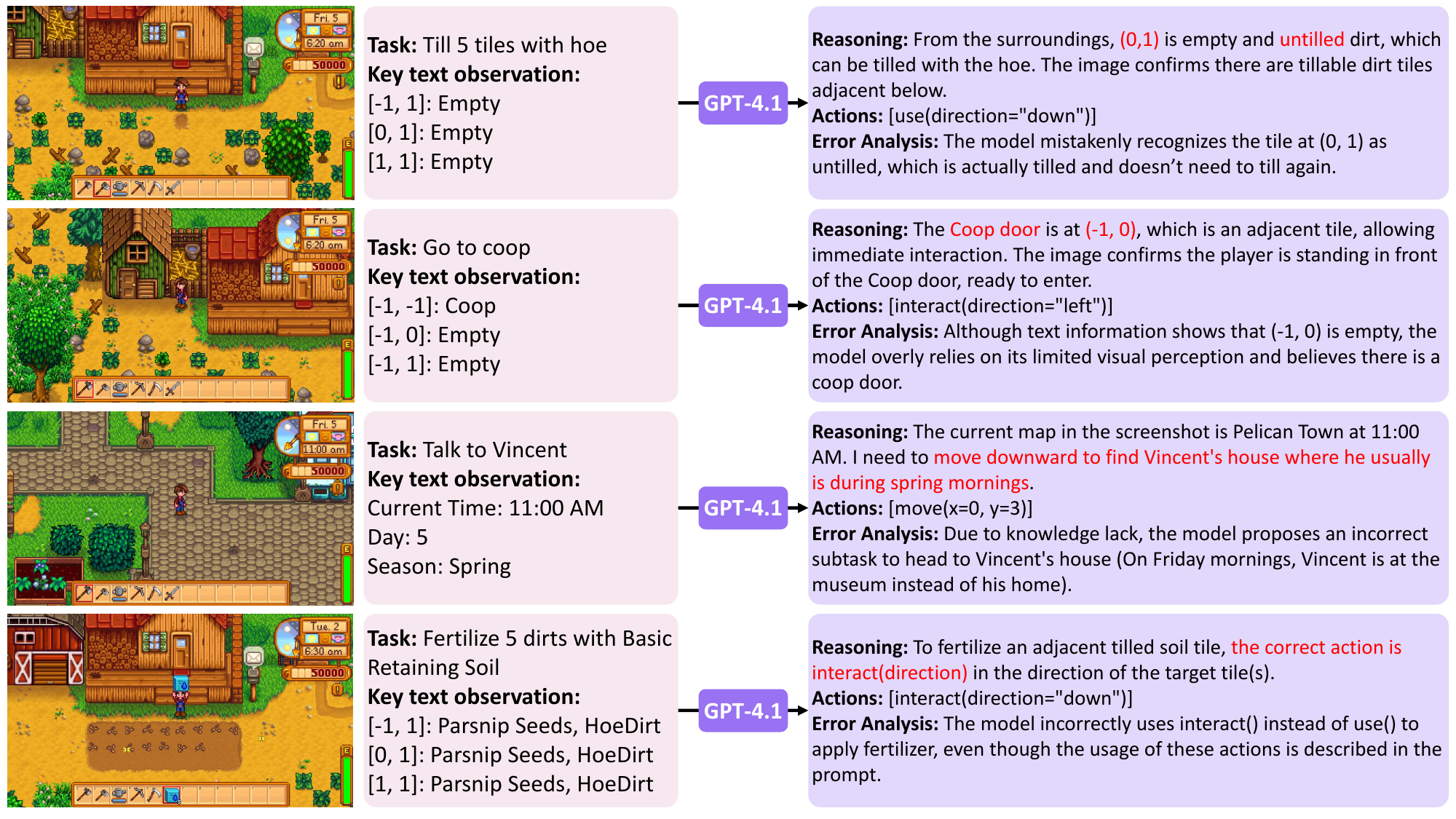}
    \caption{Cases illustrating four typical errors identified in the error analysis section: visual understanding, multimodal reasoning, long-term planning, and low-level control. For simplicity, we present only the key portions of the textual observations. Incorrect reasoning is highlighted in red, and we provide corresponding analyses to explain the causes of these errors.}
    \label{fig:four_error}
\end{figure}

In Figure~\ref{fig:four_error}, we present four representative error cases identified in the error analysis section: \textit{visual understanding}, \textit{multimodal reasoning}, \textit{long-term planning}, and \textit{low-level control}. These examples highlight common failure patterns of current agents when interacting with the environment. For clarity, we only show the key parts of the textual observations and reasoning traces.

\begin{figure}
    \centering

    \begin{subfigure}{\linewidth}
        \centering
        \includegraphics[width=\linewidth]{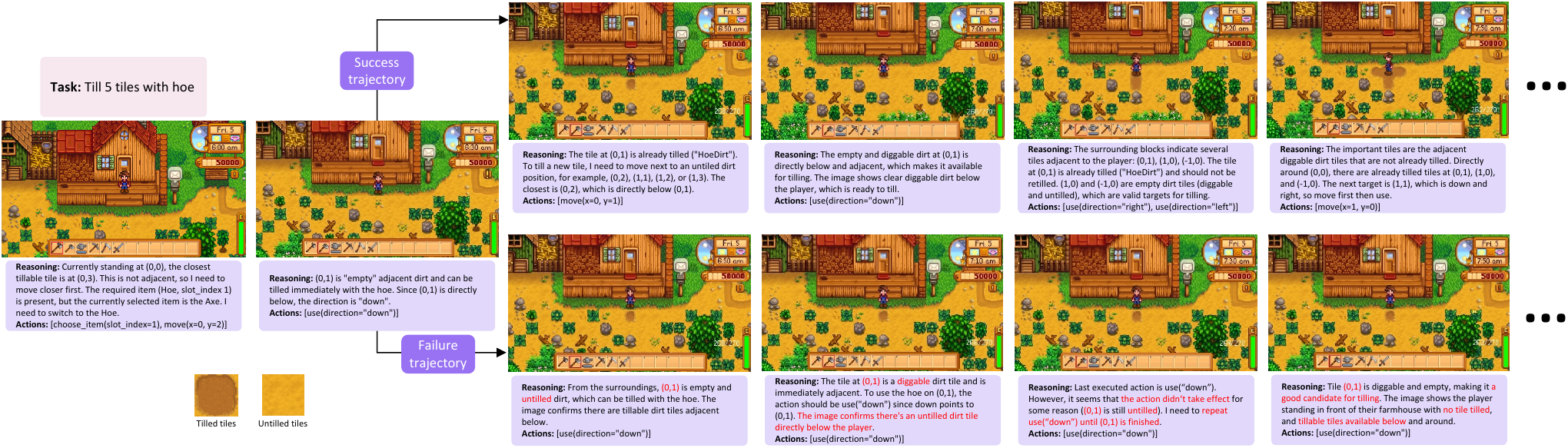}
        \caption{Task: tilling five tiles with a hoe. In the failure trajectory, the agent fails to recognize that the tile in front of the character has already been tilled, and repeatedly attempts to dig the same tile with the hoe.}
    \end{subfigure}

    \vspace{10pt}

    \begin{subfigure}{\linewidth}
        \centering
        \includegraphics[width=\linewidth]{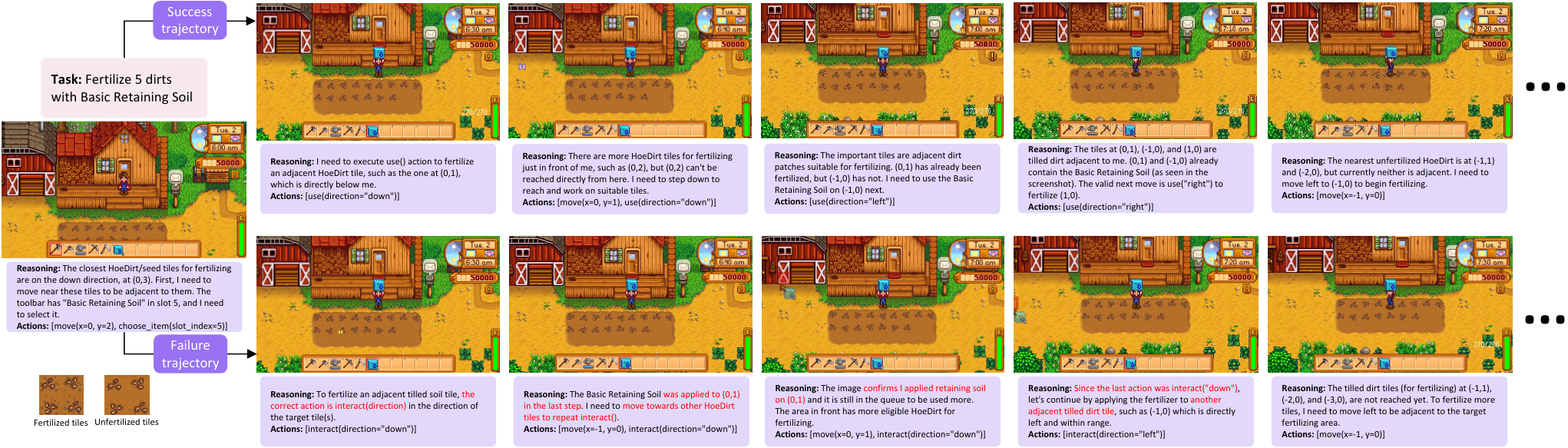}
        \caption{Task: fertilizing five dirt tiles with Basic Retaining Soil. In the failure case, although the required actions and skills are clearly explained in the prompt, the agent still executes an incorrect action. Specifically, it uses “interact” instead of “use” for fertilization. Moreover, the agent remains overconfident, incorrectly assuming that the action has been executed successfully and has taken effect.}
    \end{subfigure}

    \vspace{10pt}

    \begin{subfigure}{\linewidth}
        \centering
        \includegraphics[width=\linewidth]{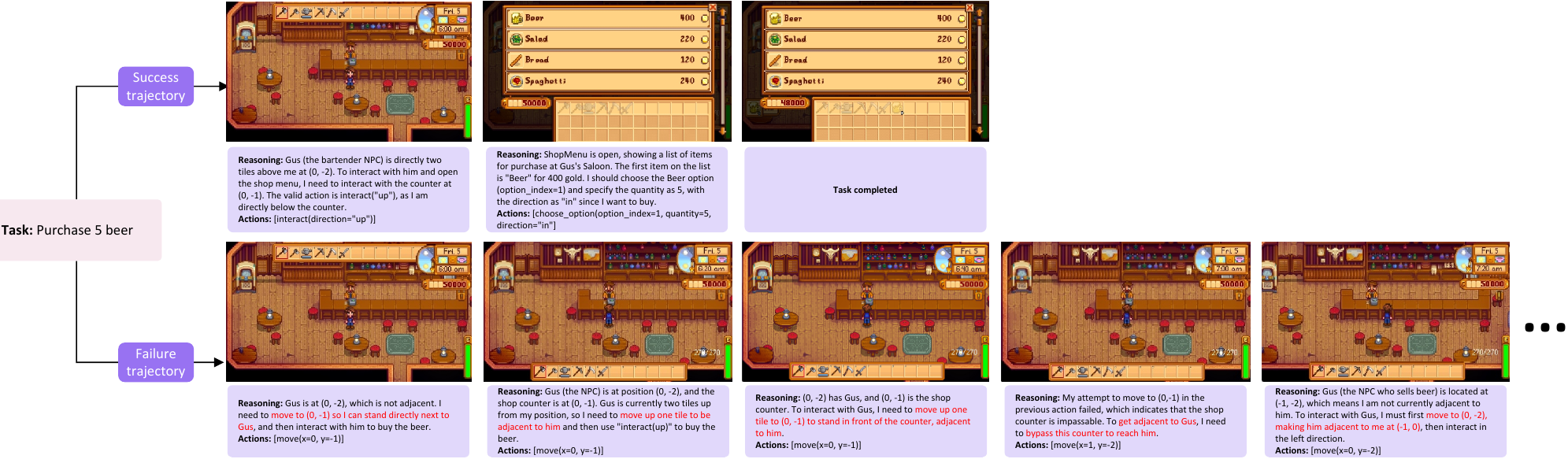}
        \caption{Task: purchasing five beers. In the successful case, the agent recognizes that the shopkeeper can be interacted with through the counter. In contrast, in the failure case, the agent attempts to move directly in front of the NPC and fails to recognize the counter as an interactable interface.}
    \end{subfigure}

    \caption{Trajectory-level case study. We present both successful and failed trajectories of a GPT-4.1 agent completing three tasks: tilling five tiles with a hoe, fertilizing five dirt tiles with Basic Retaining Soil, and purchasing five beers for comparison. Incorrect reasoning is highlighted in red.}
    \label{fig:traj_case_study}
\end{figure}

In Figure~\ref{fig:traj_case_study}, we further provide trajectory-level case studies of a GPT-4.1 agent on three representative tasks. For each task, we visualize both successful and failed trajectories to better illustrate how errors emerge during sequential decision making.

\end{document}